\def\year{2022}\relax
\documentclass[letterpaper]{article} 
\usepackage{aaai22}  
\usepackage{times}  
\usepackage{helvet}  
\usepackage{courier}  
\usepackage[hyphens]{url}  

\usepackage{graphicx} 

\usepackage{natbib}  
\usepackage{caption} 
\DeclareCaptionStyle{ruled}{labelfont=normalfont,labelsep=colon,strut=off} 
\usepackage{algorithm}
\usepackage{algorithmic}

\usepackage{newfloat}
\usepackage{listings}
\floatstyle{ruled}
\newfloat{listing}{tb}{lst}{}
\floatname{listing}{Listing}

\usepackage{booktabs}
\usepackage{amsmath}		
\usepackage{amsthm}
\usepackage{amssymb}
\usepackage{bbold}
\usepackage{mathtools}
\usepackage{enumerate}
\usepackage{pgfplots}
\pgfplotsset{compat=1.5}
\usepackage{subcaption}

\usepackage{nicefrac}
\usepackage{multirow}
\usepackage{dblfloatfix}
\usepackage{subcaption}
\usepackage{mathrsfs}

\newcommand{\R}{\mathbb{R}}

\newcommand{\VV}{\mathcal{V}}
\newcommand{\WW}{\mathcal{W}}
\newcommand{\XX}{\mathcal{X}}

\makeatletter
\newcommand{\printfnsymbol}[1]{%
    \textsuperscript{\@fnsymbol{#1}}%
}
\makeatother

\nocopyright
\usepackage{hyperref} 

\title{Explainability Requires Interactivity}
\author{
    Matthias Kirchler \thanks{The first two authors contributed equally to this work}\textsuperscript{\rm 1,2},    
    Martin Graf \printfnsymbol{1}\textsuperscript{\rm 1},
    Marius Kloft \textsuperscript{\rm 2},
    Christoph Lippert \textsuperscript{\rm 1,3}
}
\affiliations{
    \textsuperscript{\rm 1} Hasso Plattner Institute for Digital Engineering, University of Potsdam, Germany\\
     \textsuperscript{\rm 2} TU Kaiserslautern, Kaiserslautern, Germany\\
     \textsuperscript{\rm 3} Hasso Plattner Institute for Digital Health, Icahn School of Medicine at Mount Sinai, New York, NY 10029\\
    \{matthias.kirchler,christoph.lippert\}@hpi.de, martin.graf@student.hpi.uni-potsdam.de, kloft@cs.uni-kl.de
}

\begin{document}

\maketitle

\begin{abstract}
When explaining the decisions of deep neural networks, simple stories are tempting but dangerous.
Especially in computer vision, the most popular explanation approaches give a false sense of comprehension to its users and provide an overly simplistic picture.
We introduce an interactive framework to understand the highly complex decision boundaries of modern vision models.
It allows the user to exhaustively inspect, probe, and test a network's decisions.
Across a range of case studies, we compare the power of our interactive approach to static explanation methods, showing how these can lead a user astray, with potentially severe consequences.\footnote{We release the interactive framework under \url{https://github.com/HealthML/StyleGAN2-Hypotheses-Explorer}, while code to reproduce results is published under \url{https://github.com/HealthML/explainability-requires-interactivity}}
\end{abstract}

\begin{figure}[t!]
    \centering
    \begin{subfigure}[b]{0.49\textwidth}
        \centering
        \includegraphics[clip,trim={13px, 14px, 10px, 0},width=\textwidth]{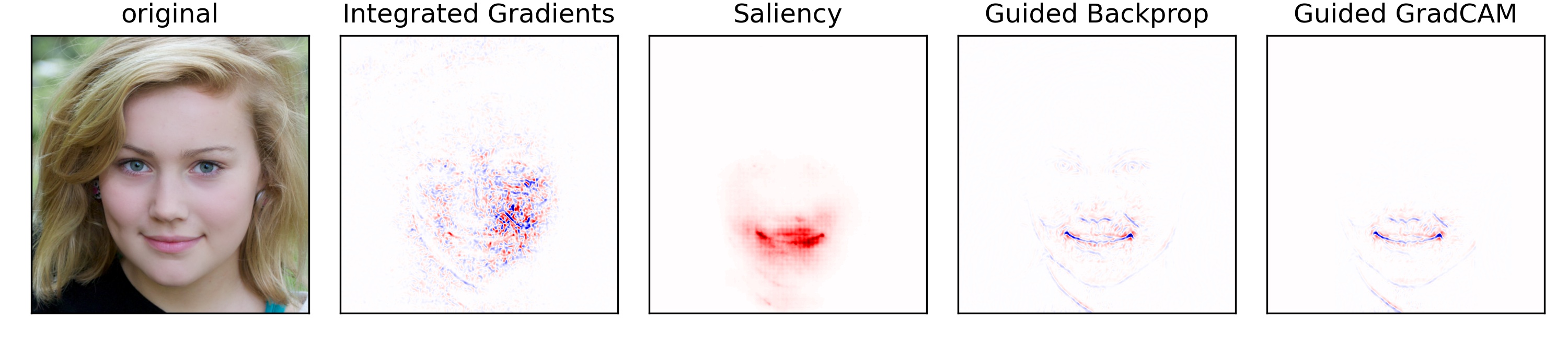}
        \caption{Heat map methods.}
    \end{subfigure}
    \begin{subfigure}[b]{0.49\textwidth}
        \centering
        \includegraphics[ width=\textwidth]{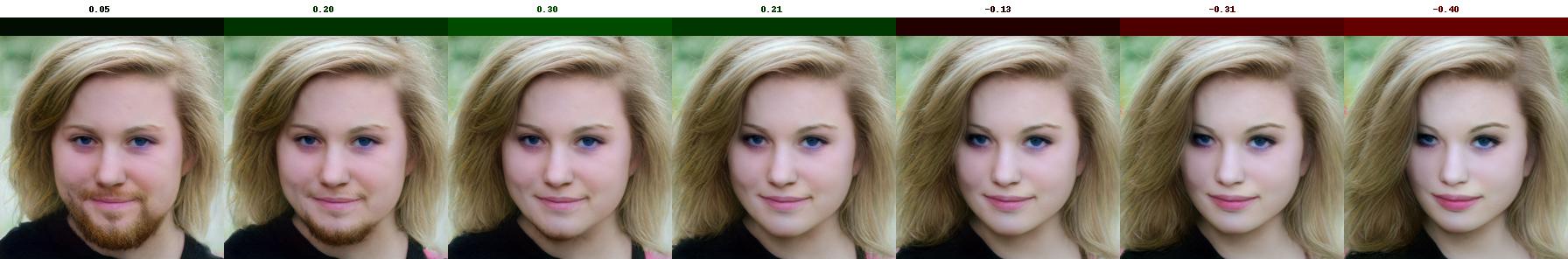}
        \caption{Automated (static) change of ``makeup'' variable.}
    \end{subfigure}
    \begin{subfigure}[b]{0.49\textwidth}
        \centering
        \includegraphics[width=\textwidth]{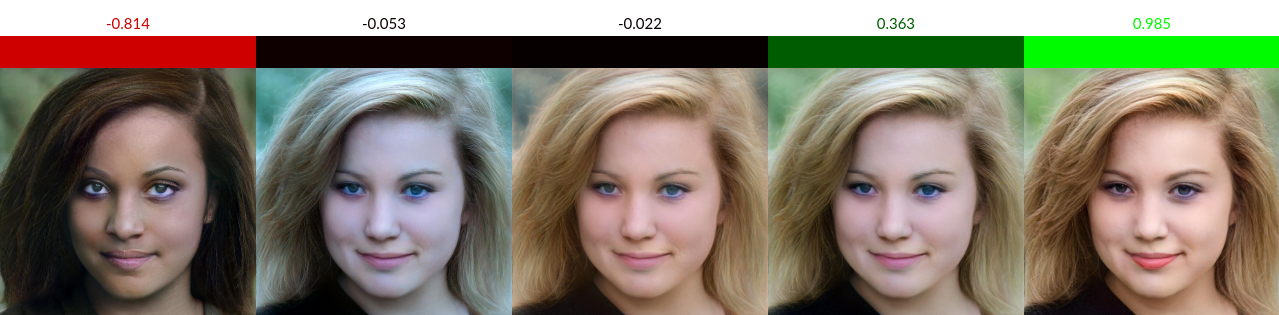}
        \caption{Snapshot of manual, interactive modifications (ours).}
    \end{subfigure}    
    \caption{
        Static approaches (a and b) and a snapshot of our interactive approach (c) for a smile classifier (1=``smiling'', -1=``not smiling''):
        The interactive approach detects features such as skin tone, hair color, or the amount of makeup as strong influences in the classification of the network.
        The attention heat maps almost exclusively focus on the region around the mouth; the automated generative approach accidentally changes confounding factors such face morphology and facial expression.
    }
    \label{fig:teaser}
\end{figure}
With the improved power and increasing number of applications of deep neural networks in real-world settings, practitioners need to evaluate, understand, and explain their models.
In recent years, explainable AI (XAI) has been one of the most active research areas in computer vision and beyond.

The most prominent strain of explanation methods in vision applications consists of constructing heat maps that illustrate where a neural network is focusing its attention in an image, such as in saliency maps \citep{simonyan2013deep} and class activation maps \citep{selvaraju2017grad}.
Other explanation methods include visualizing individual convolutional filters by optimizing the filters' output over the input image \citep{olah2017feature} or comparing images to reference prototype images \citep{chen2019looks}.

Common to almost all current interpretation approaches is that they only provide \emph{static} explanations.
However, static explanations, and especially heat map approaches, are problematic for several reasons.
First, they are vulnerable to confirmation bias: it is easy to find and interpret ostensible patterns in heat maps that do not reflect the actual decision process of the network.
For an example, consider Figure~\ref{fig:teaser}, showing a comparison of explanation methods for a smile classification network.
A human might expect the network to focus on the nasolabial region and especially the mouth.
The heat map visualizations in (a) appear to confirm this, as they direct almost all their attention to this area.
However, by interactively exploring variations of the face (c), it can be seen that the classifier reacts extremely sensitive towards unrelated factors such as skin color (darker skin leads to a ``not smiling'' classification) or the level of makeup (more makeup leads to a ``smiling'' classification).
This effect is present even when keeping the facial features and mouth expression virtually constant.
This shortcoming may inadvertently reinforce systemic biases such as racism and sexism and can be dangerous in safety-critical applications.
E.g., in medical diagnostics or autonomous driving, a practitioner might gain a false sense of comprehension after studying only the static explanations but remains oblivious to potentially fatal shortcomings of the model.
Second, heat maps can only provide a minimal glance into the decision-making process of a neural network.
In the example from Figure~\ref{fig:teaser}, it is not clear what about and how the focused area would need to change for the model to classify the image differently.
In contrast, an interactive explanation method allows exploring multiple variations of the face, such as different levels of makeup, skin tones and hair colors, and facial expressions.
Finally, another troublesome aspect of heat map approaches is that while the methods technically can detect and react to color information, they are ill-suited to communicate the model's behavior concerning color. 
Consequently, explanations remain ``color-blind'' even and especially in situations where awareness of color information is crucial to detect biases, such as for smile detection, age regression, or perceived gender classification.

While some static explanation methods automatically create variations of an input image to provide explanations, they face their own limitations; e.g., in Figure~\ref{fig:teaser}-b, the image is automatically changed along a learned ``makeup'' dimension that was trained on a large number of labeled examples.
However, the direction is confounded both with face morphology features and facial hair, and may potentially introduce image artifacts.
Generally, quality control in these methods is challenging, and it is even harder to counteract artifacts and confounding factors.
Even if a direction provides satisfactory results, it only provides a limited glance into the network, without the option for broader exploration.
In contrast, in an interactive approach, each image can be modified to faithfully encode the user's goals.
The model can then be iteratively probed to explore its decision boundary exhaustively.

We argue that the analysis of black-box computer vision predictors can be enhanced considerably by including an \emph{interactive} component.
Only very few works so far have investigated how interactive elements can be used to better understand both a classifier's internal decisions and dataset-specific artifacts.
By leveraging recent advances in high-quality image generation and editing, we present a framework that allows users to fluidly modify images and interactively explore the classification decisions of a neural network.
In particular, a user can \emph{(i)} form a hypothesis as to what input features a predictor is sensitive towards, then \emph{(ii)} encode these features through the interface, and finally \emph{(iii)} get a verification or falsification of this hypothesis.

Our framework enables two critical applications: 1) it allows users to learn interactively what a neural network pays attention to; and 2) it allows to explore the underlying data distribution and thus to potentially uncover both novel scientific facts and confounding ``Clever Hans'' artifacts \citep{lapuschkin2019unmasking}.

\section{Related Work}
Much prior work has been dedicated to the explainability of deep learning and vision models; see, e.g., \citep{vilone2020explainable} for a recent systematic review.

\paragraph{Heat Map Methods}
Many explanation approaches generate heat maps that localize a neural network's attention within an image \citep{bach2015pixel,fong2017interpretable,dabkowski2017real,selvaraju2017grad,shrikumar2017learning,simonyan2014deep,zeiler2014visualizing}.
These methods can be helpful in certain situations, e.g., when a classifier focuses on confounding factors, such as watermarks or copyright notices embedded within images instead of the real class-defining features \citep{lapuschkin2019unmasking}.
Besides the shortcomings of heat map methods mentioned in the introduction, it has been shown that many saliency methods produce unreliable and misleading explanations that often are not better than entirely random feature attribution \citep{hooker2019benchmark,adebayo2018sanity,kindermans2019reliability,yang2019benchmarking}.
Generally, heat map attribution methods can be powerful in \emph{localizing} the attention of a network but fall short of actually \emph{explaining} the network's decisions.

\paragraph{Automated Generative Methods}
Some explanation methods have been proposed that use generative models.
\citet{burns2020interpreting} automatically modify images while controlling for ``false positive'' explanations; \citet{singla2019explanation,denton2019image,liu2019generative,li2021discover} propose different variations of modifying images in the code space of a generator network and move along or across decision surfaces of classifiers or regressors.
\citet{balakrishnan2020towards,goyal2019counterfactual,joshi2019towards} leverage causal frameworks to discover biases and interpret image modifications as counterfactuals.
Many of these methods are geared towards face analysis systems and are non-trivial to generalize to other settings.
While automated generative explanation methods overcome many of the problems of  heat map approaches, they exhibit a new set of potential problems.
The automatically generated directions in code space in these works, both unsupervised and supervised, require a high level of disentanglement of the generator's code space, which in practice is not guaranteed, especially considering the impossibility results for strong disentanglement in unsupervised settings \citep{locatello2019challenging}.
Automated methods can also lead to unwanted image artifacts due to insufficient quality control, considerably impairing a method's viability.
Both failure modes can be hard to detect and even harder to counteract in practice.
In contrast, in an interactive approach, a user can decide on each instance if only the features of interest are modified and whether unnatural artifacts have been introduced.  
Several methods also require large-scale annotations for all features to be investigated (e.g., \citet{balakrishnan2020towards} report 0.52M manual annotations and other methods use the large set of CelebA labels).
This labeling can be costly and time-intensive, and it is hard to guarantee a high quality of labels.
In domains such as medical diagnostics, annotations at this scale are infeasible due to the required expertise for the annotators and the considerably longer time each annotation can take.
Our interactive framework does not require any annotations.

\paragraph{Interactive Approaches}
While natural language models can be explored interactively by modifications of the input data, e.g., replacing named or other entities within a sentence \citep{prabhakaran2019perturbation,garg2019counterfactual}, images are considerably harder to modify realistically.
Some works have introduced interactivity into the analysis of vision models \citep{olah2018the,li2020visualizing,carter2019activation} but do not allow modification of images.
Closest in spirit to ours is the work by \citet{cabrera2018interactive}, which allows users to interactively remove parts of an input image that get in-painted by classical computer vision algorithms.
We believe that this approach can be greatly enhanced by using a wider array of image modifications, using modern generative deep learning models.

\section{Methods}
In this section, we explain our methodology to explore a neural network's decision interactively.

Given an input space of images, $\XX \subset \R^{h \times w \times 3}$, we consider a setting with a trained binary classifier $f: \XX \to [-1, 1]$, where a large score $f(x)$ indicates higher confidence in classifying the image $x$ into class $1$, and vice versa.
The framework is easily extended to regression and multi-class settings.
In contrast to other data modalities such as text and tabular data, it is more challenging to alter images authentically without breaking away from the manifold of realistic images.
Therefore, we utilize a generative deep learning model $g: \VV \to \XX$ that maps a latent code $v \in \VV$ to an image $g(v)$.
Instead of directly altering the image $x \in \XX$, a user only directly interacts with and modifies its corresponding latent code $v$.

We consider three main desiderata for choosing a generative model for interactive exploration: 1) quality of the generated images; 2) disentanglement of the latent code space $\VV$; and 3) expressivity of the model, i.e., whether the generator can cover the whole space of relevant input images.
Different forms of representing and manipulating the latent code will be appropriate, depending on which kind of generator architecture $g$ is chosen (e.g., different variants of variational autoencoders \citep{kingma2013auto} and generative adversarial networks \citep{goodfellow2014generative}).
We focus on using a StyleGAN2 \citep{karras2020analyzing} as generator, but the methodology can be easily adapted for other architectures.

Prior work has shown that StyleGAN2 both achieves high image quality \citep{karras2020analyzing} and has a well disentangled latent space \citep{wu2021stylespace}.
One drawback of the StyleGAN2 architecture is that it can still exhibit considerable mode collapse \citep{yu2020inclusive}; however, we can circumvent this problem as described later, since images can still be inverted faithfully \citep{karras2020analyzing,wu2021stylespace}

The StyleGAN2 architecture generates images by first mapping a standard normal random vector $z \in \R^d$ to a ``style'' vector $w = h(z) \in \WW \subset \R^{d'}$ with a fully connected neural network $h$.
The style vector $w$ is then fed into each of the $L$ layers of the generator network to generate an image.
Different styles can be mixed by using different style vectors $w_1, \ldots, w_L$ in each layer.
For a given real image, a corresponding input style $w$ can be found by minimizing a reconstruction error over the input style via descent methods \citep{zhu2016generative}.
\citet{wu2021stylespace} have shown that superior disentanglement and reconstruction can be achieved by using the so-called ``StyleSpace,'' a downstream layer-wise transformation of $w$; however, this improvement considerably limits the editing capabilities of reconstructed images.
Improved reconstruction can also be realized by optimizing over different styles at different layers, but this again comes at a cost to editing capabilities.
As a trade-off, we found it satisfactory to reconstruct images by optimizing over shared styles $w$ in each layer.
For editing images, we mix different styles $w_1, \ldots, w_L$ at different layers.

We developed an interface\footnote{
The interface is loosely inspired by the presentation of \citet{karras2020analyzing} at \url{https://www.youtube.com/watch?v=9QuDh3W3lOY} for a different application.
}  (see Figure~\ref{fig:overview} in the Appendix) in which users can interactively mix different styles $w_1, \ldots, w_L$ for different layers to generate and modify images.
The styles can be either randomly generated or can be the reconstruction of user-input images.
The generator creates a fixed number of output images from these styles.
The interface orders the output images by classification score and presents the score on top of the image.
Users can then edit those output images to gain an understanding of the behavior of the classification network.
There is a fundamental trade-off to be made between the expressiveness of an interface and its usability.
In our case, in each layer, a user can either select a single style or linearly interpolate between two styles.
Adjacent layers can be grouped since otherwise, the user needs to adjust up to 18 different styles (for images of size $h = w = 1024$) for a single modification.
Each style can be represented either by the result if that style was selected for this layer (``result view'') or by the corresponding image was the style fed into all layers at the same time (``style view'').

Next, we discuss two critical application scenarios and how an interface such as ours can enhance the interpretability of the classification network.

\paragraph{Application Scenario 1: Understanding Individual Decisions}
\label{sec-exp}
\begin{figure*}[t]
    \centering
    \begin{subfigure}[b]{0.49\textwidth}
        \centering
        \includegraphics[clip,trim={13px, 14px, 10px, 0},width=\textwidth]{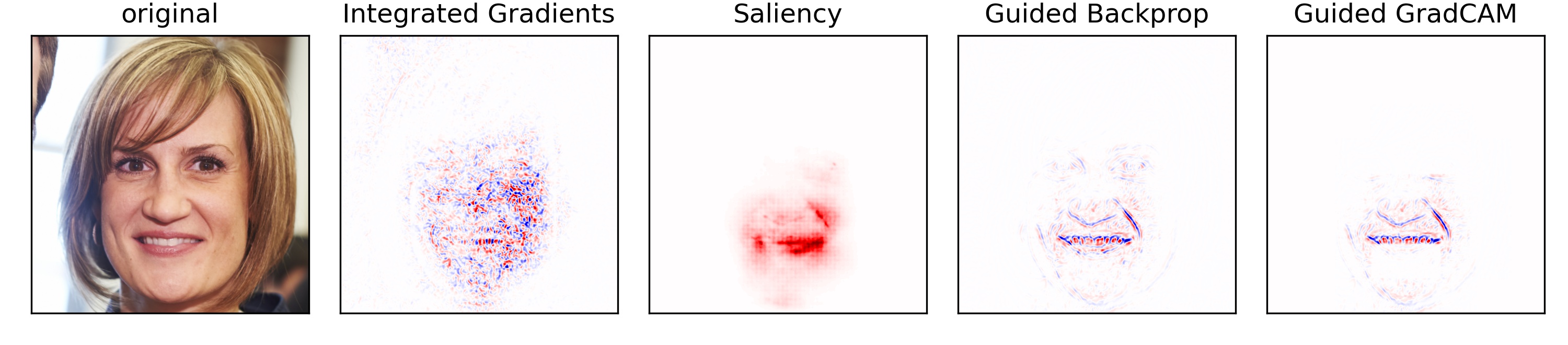}
    \end{subfigure}
    \begin{subfigure}[b]{0.49\textwidth}
        \centering
        \includegraphics[clip,trim={13px, 14px, 10px, 0},width=\textwidth]{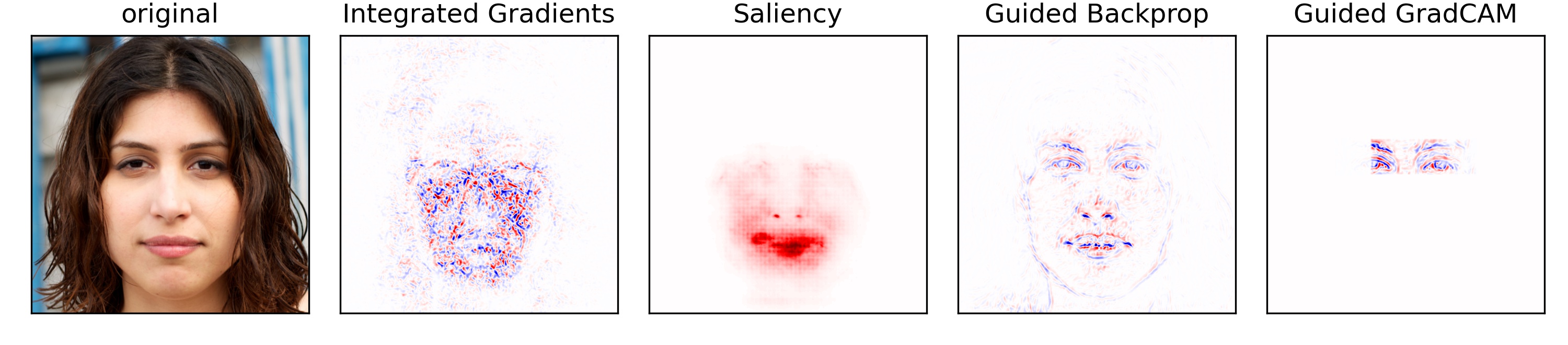}
    \end{subfigure}
    \begin{subfigure}[b]{0.49\textwidth}
    \centering
        \includegraphics[width=\textwidth]{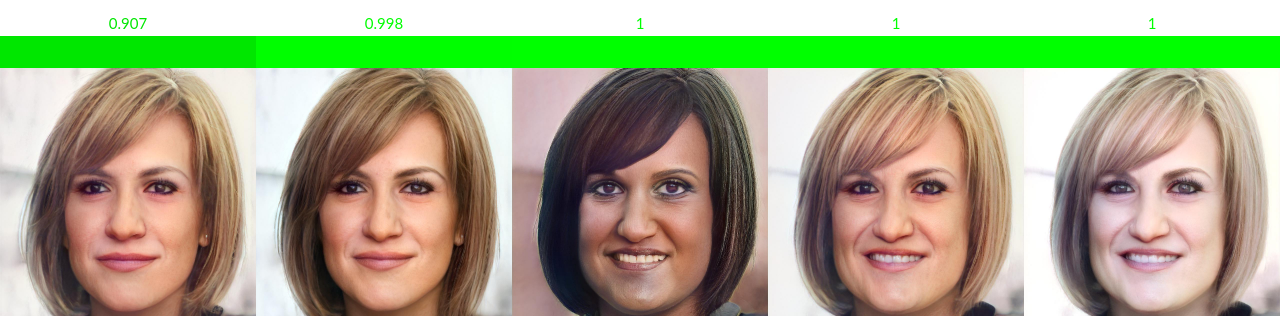}
        \caption{Individual 20053}
    \end{subfigure}
    \begin{subfigure}[b]{0.49\textwidth}
        \centering
        \includegraphics[width=\textwidth]{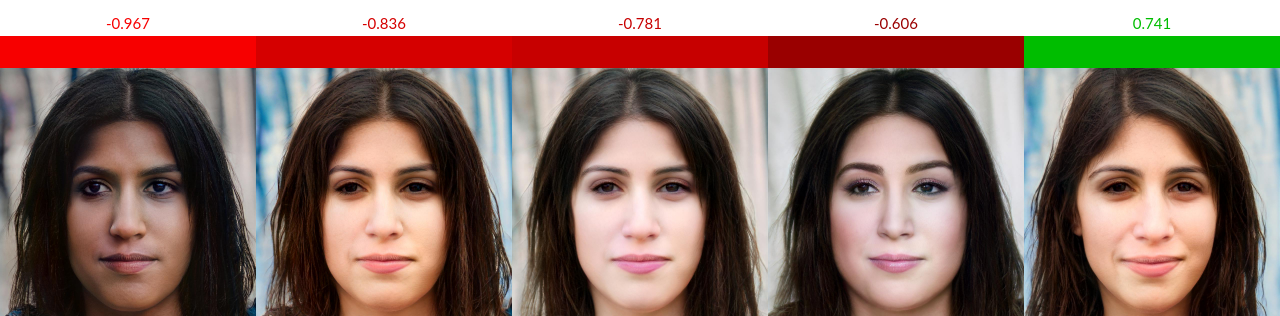}
        \caption{Individual 20827}
    \end{subfigure}
    \begin{subfigure}[b]{0.49\textwidth}
        \centering
        \includegraphics[clip,trim={13px, 14px, 10px, 0},width=\textwidth]{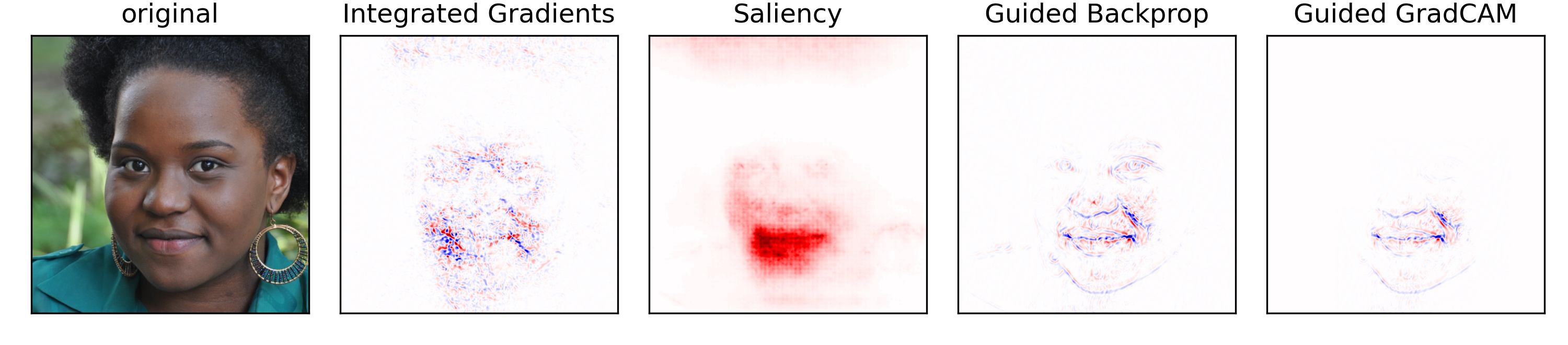}
    \end{subfigure}
    \begin{subfigure}[b]{0.49\textwidth}
        \centering
        \includegraphics[clip,trim={13px, 14px, 10px, 0},width=\textwidth]{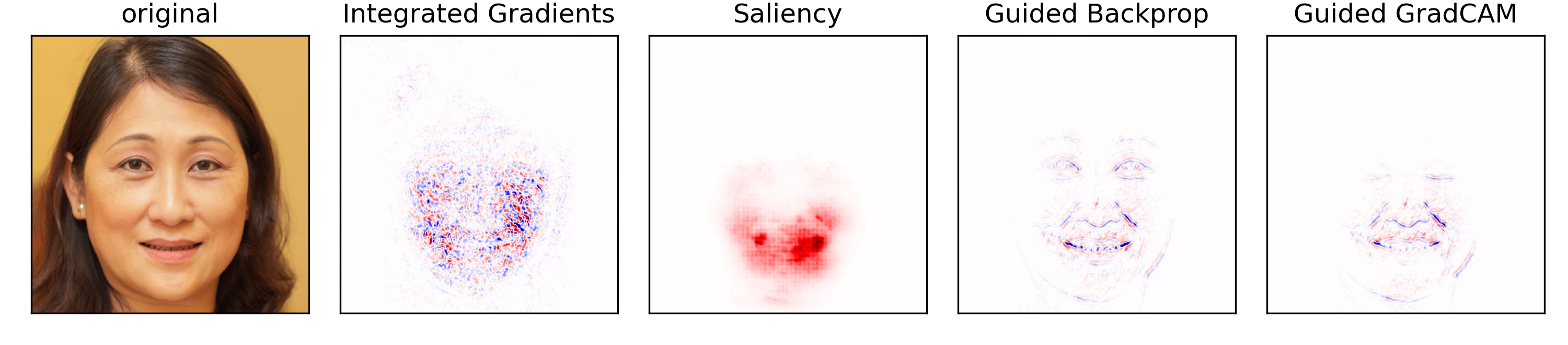}
    \end{subfigure}
    \begin{subfigure}[b]{0.49\textwidth}
    \centering
        \includegraphics[width=\textwidth]{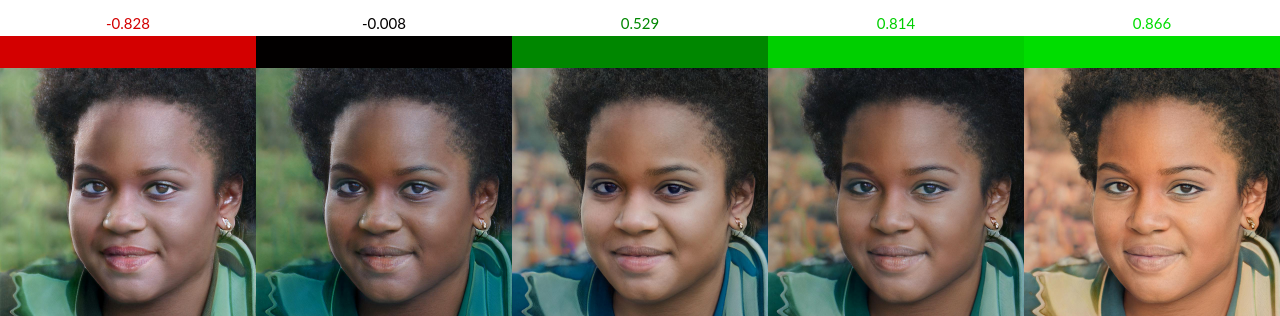}
        \caption{Individual 20659}
    \end{subfigure}
    \begin{subfigure}[b]{0.49\textwidth}
        \centering
        \includegraphics[width=\textwidth]{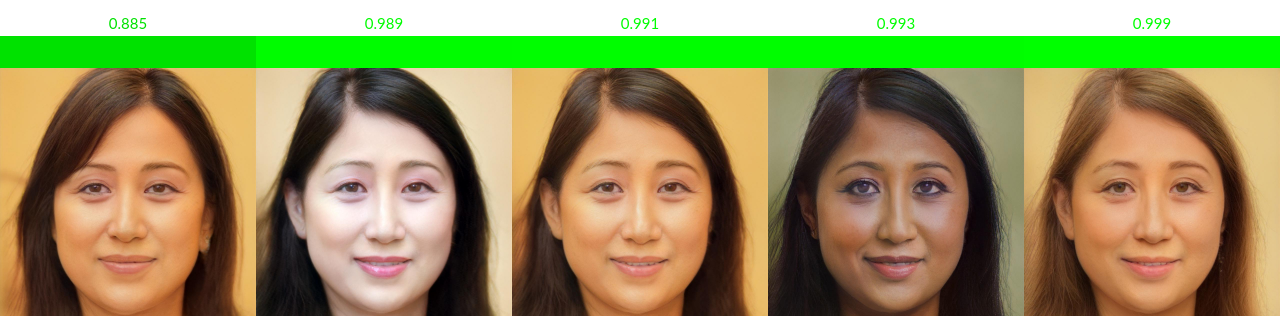}
        \caption{Individual 20000}
    \end{subfigure}
    \caption{
    Smile Classification.
    Comparison of interactive and static explanations on four different images in FFHQ (ids indicate filename in dataset).
    In each top row are the original image and four different heat map visualizations.
    In each second row, a snapshot of our interactive visualization is displayed; images are sorted by classification score (shown on top of each image).
    Negative (red) class is ``not smiling'', positive class (green) is ``smiling''.
    }
    \label{fig:smile}
\end{figure*}
\begin{figure*}[t]
    \centering
    \begin{subfigure}[b]{0.49\textwidth}
        \centering
        \includegraphics[clip,trim={13px, 14px, 10px, 0},width=\textwidth]{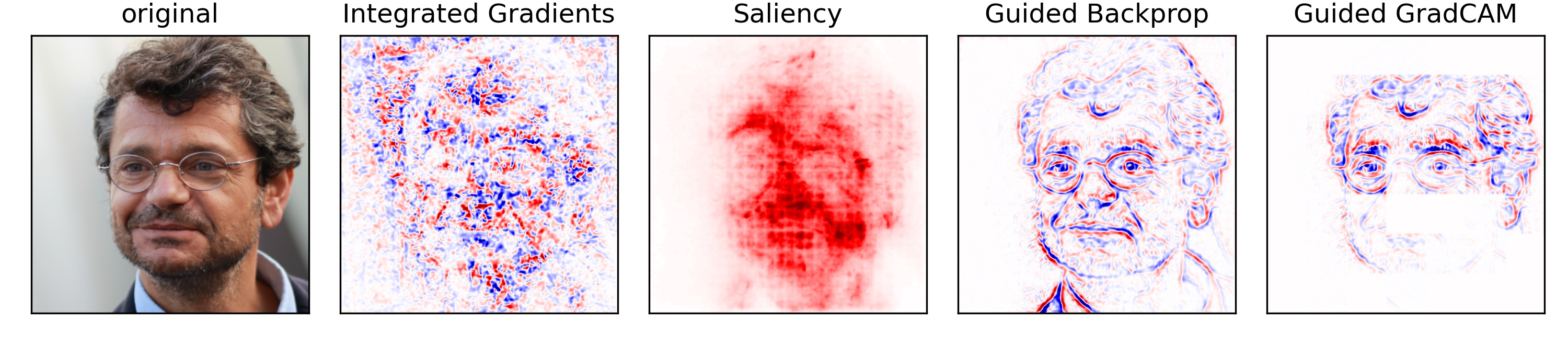}
    \end{subfigure}
    \begin{subfigure}[b]{0.49\textwidth}
        \centering
        \includegraphics[clip,trim={13px, 14px, 10px, 0},width=\textwidth]{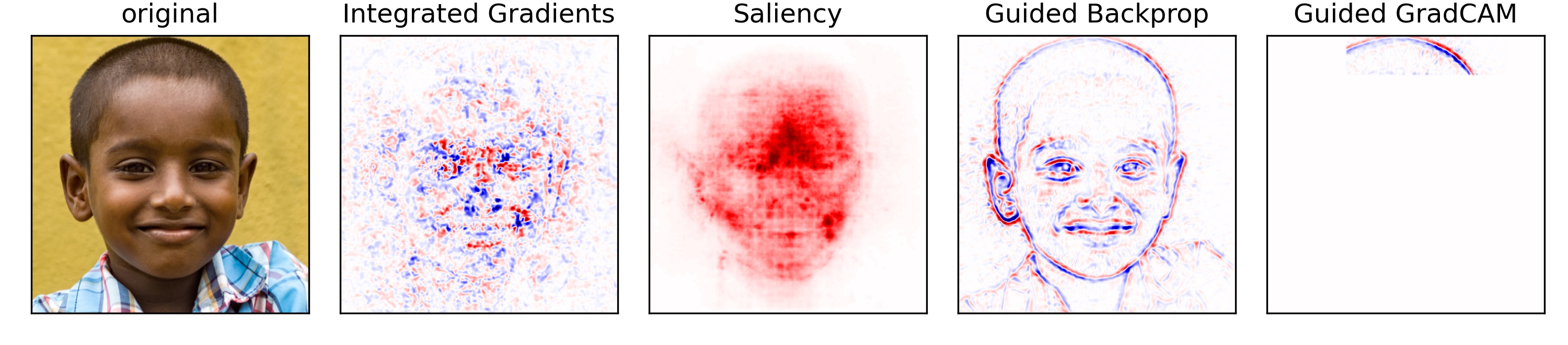}
    \end{subfigure}
    \begin{subfigure}[b]{0.49\textwidth}
    \centering
        \includegraphics[width=\textwidth]{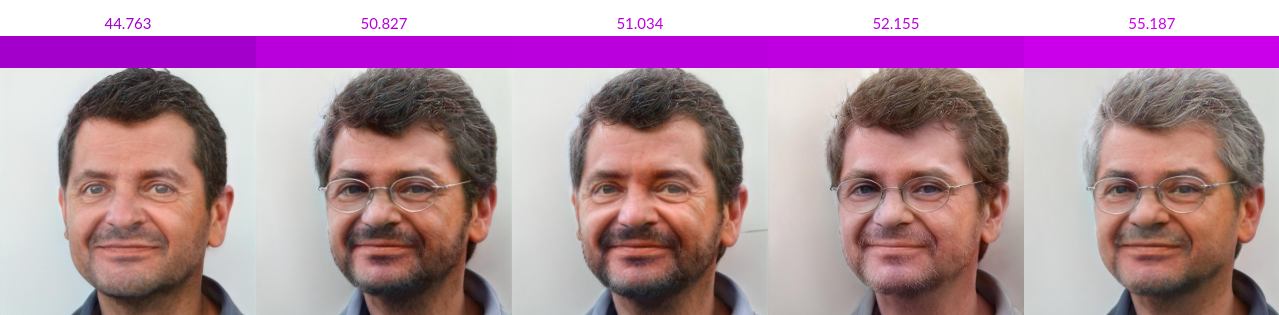}
        \caption{Individual 30952}
    \end{subfigure}
    \begin{subfigure}[b]{0.49\textwidth}
        \centering
        \includegraphics[width=\textwidth]{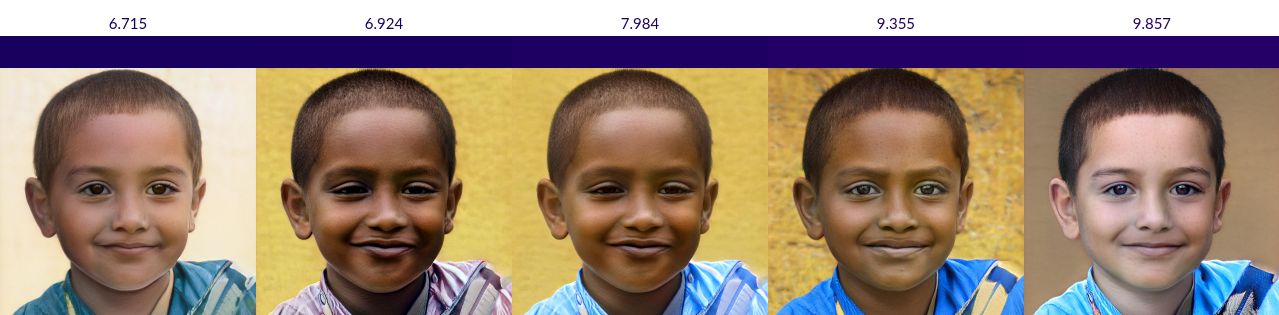}
        \caption{Individual 40923}
    \end{subfigure}
    \begin{subfigure}[b]{0.49\textwidth}
        \centering
        \includegraphics[clip,trim={13px, 14px, 10px, 0},width=\textwidth]{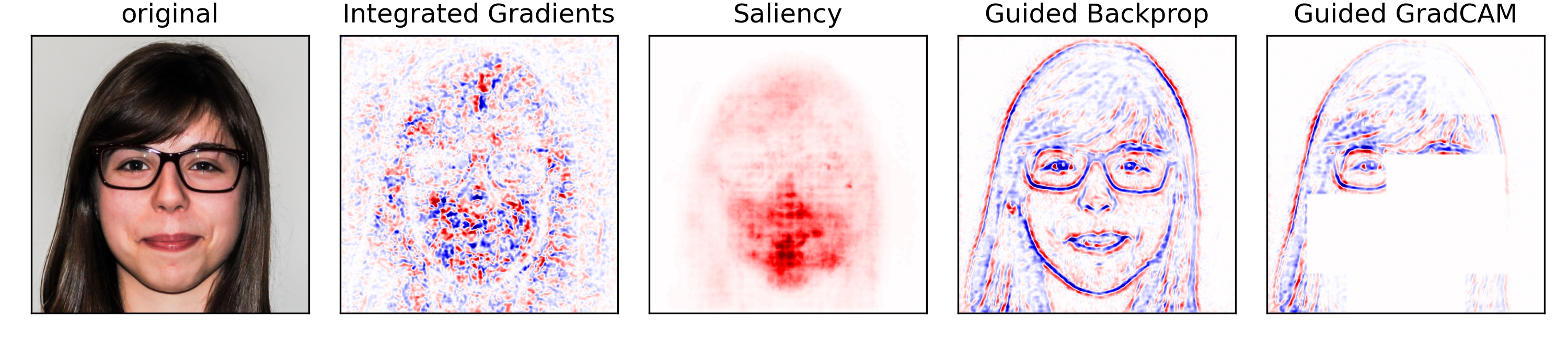}
    \end{subfigure}
    \begin{subfigure}[b]{0.49\textwidth}
        \centering
        \includegraphics[clip,trim={13px, 14px, 10px, 0},width=\textwidth]{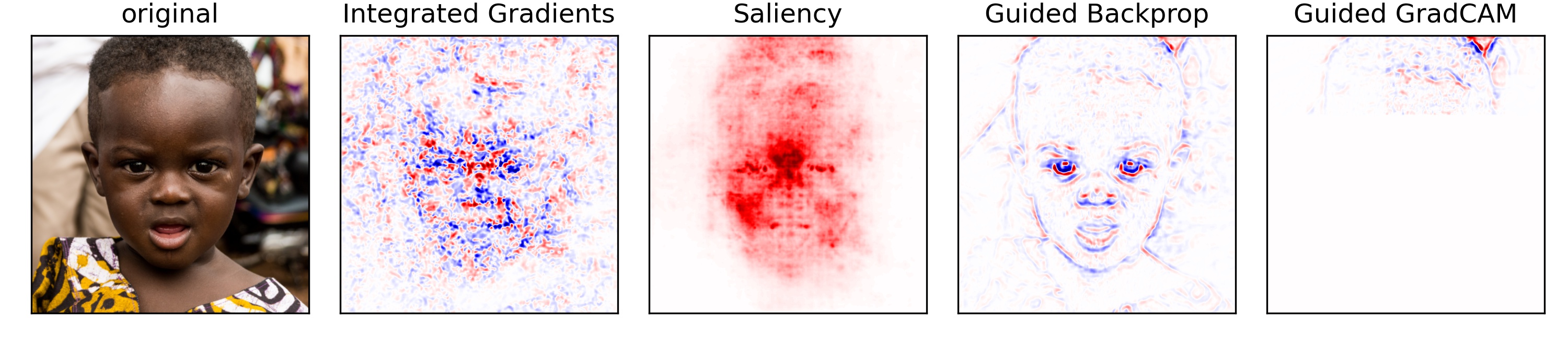}
    \end{subfigure}
    \begin{subfigure}[b]{0.49\textwidth}
    \centering
        \includegraphics[width=\textwidth]{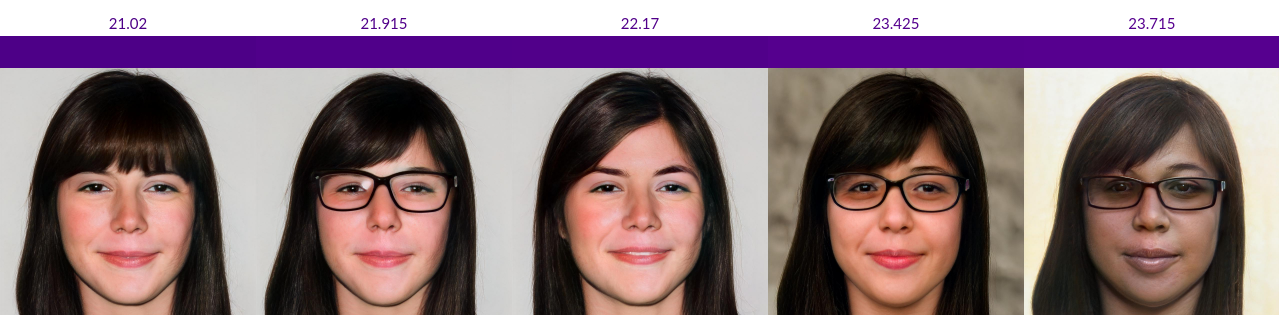}
        \caption{Individual 40151}
    \end{subfigure}
    \begin{subfigure}[b]{0.49\textwidth}
        \centering
        \includegraphics[width=\textwidth]{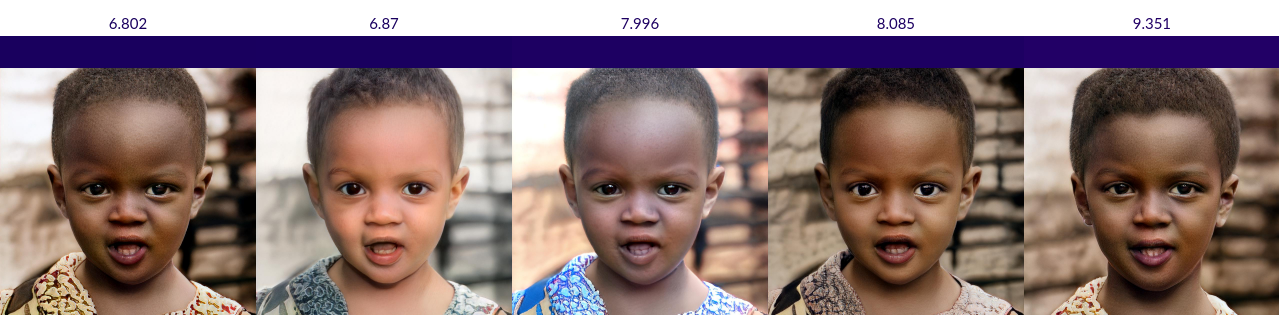}
        \caption{Individual 40895}
    \end{subfigure}
    \caption{
    Age Regression.
    Comparison of interactive and static explanations on four different images in FFHQ.
    }
    \label{fig:age}
\end{figure*}

The main scenario for ML interpretability is to understand why an image classifier makes a particular decision on a specific image.
Safety-critical applications include, for example, for a physician to understand why a medical image was classified as healthy or diseased; or for an autonomous driving engineer to understand why an image was classified as either bicycle or motorcycle.
Standard approaches are primarily based on heat maps, with few exceptions based on automatically generating counterfactual images or comparisons to prototypes (see the section on related work).

In our proposed framework, the generative model inverts the image under consideration to find the corresponding latent code $w$.
An additional set of independent latent codes can be created from reference images containing high-level features the user wants to investigate.
The selection of reference images strongly influences the quality of mixed images, and which aspects of the classifier can be inspected.
We found the most promising results when selecting reference images similar along dimensions that we do not want to explore but dissimilar in a few key aspects of interest.
For example, in a smile detection setting, we select reference images of the same perceived gender, similar age, and similar face orientation, but potentially different skin tone \& ancestry or different in makeup usage.
However, if we want to probe different facets of the classifier, we need to select different reference images.
Alternatively or additionally, randomly generated latent codes can be used for image modification.

While copies of the reconstructed image under investigation can also be automatically optimized to move towards the opposite ends of the classification spectrum (similar to \cite{denton2019image,schutte2021using}) we found that this usually degraded image quality too much without a clear advantage to hand-selected alternate images.


\paragraph{Application Scenario 2: Understanding Data Distributions and Models}
A more exploratory scenario comes up when we want to understand more broadly what kind of patterns a neural network has found in a data set.
This is especially important to find confounding factors in the data that may not generalize to a network's deployment settings, such as in the case of the ``Clever Hans'' effect; for example, a network may notice that images of horses tend to contain copyright statements \citep{lapuschkin2019unmasking}, but if the trained network is to be deployed on images without these factors, it will likely suffer from a performance drop and exhibit counter-intuitive behavior.

Another use case lies in basic science research.
A non-parametric hypothesis test on images \citep{lopez2017revisiting,kirchler2020two} may have found a difference between two sets of images.
One group of images (class $+1$) may be medical images (such as retinal fundus images or cardiac MRI) of patients with a specific genetic mutation.
The other group of images (class $-1$) may be images of patients without the genetic mutation.
While non-parametric hypothesis tests can be powerful in finding these differences and guarantee high statistical significance, they usually do not provide insights for a human to understand the exact differences between the two groups.
In contrast to many standard machine learning problems in computer vision, a domain expert does \textbf{not} know the difference between the classes beforehand, and the effect may be very subtle.

Understanding a network's decision in a general sense and fully understanding the complex underlying image distribution are inherently linked.
In our framework, a user can develop a novel hypothesis about the difference between the two classes, then encode this hypothesis with the interactive interface and get feedback from the classification network, essentially verifying or falsifying the hypothesis.

\section{Experiments}
This section investigates three different prediction models on facial images: smile classification, age estimation, and perceived gender classification.
In all three settings, we use a StyleGAN2 trained on the FFHQ dataset and evaluate the corresponding predictors on images also selected from FFHQ.

We compare against the baseline methods Integrated Gradients \citep{sundararajan2017axiomatic}, Saliency Maps (with noise tunnel, \citep{simonyan2013deep, smilkov2017smoothgrad}), Guided Backprop \citep{springenberg2014striving}, and Guided GradCAM \citep{selvaraju2017grad}, all implemented in the Captum library \citep{kokhlikyan2020captum}, as well as automated generative explanation methods, similar to \citep{denton2019image,schutte2021using}.

In addition to exploring single decisions on facial images, we show how a more explorative approach can yield insights into a classifier's inner workings on a flower classification task.

\subsection{Smile Detection}
We first consider a ResNet18 \citep{he2016deep} trained on CelebA-HQ to distinguish smiling against not smiling faces \citep{liu2015deep,karras2017progressive,lee2020maskgan}.
Figures~\ref{fig:teaser} and ~\ref{fig:smile} show the comparisons.
For all five images, the attribution methods focus mainly on the region around the mouth and partially around the eyes.
This intuitive result is broadly in line with how humans tend to recognize smiles \citep{frank1993not}.
Note that the individual heat map methods do not fully conform with one another.
Especially the Integrated Gradients tend to cover a wider range, including the nasolabial region and eyes and also broader regions around the cheeks.
However, it is unlikely that this indicates skin color since other exposed skin regions (such as the forehead and neck) are not covered.

We investigated the same images in the interactive approach and recorded a snapshot of modified images  with classification scores.
We first note that in  Figure~\ref{fig:smile}-a (the two leftmost images) and d (leftmost image), a direct modification of the mouth from open to closed leads to only a minimal drop in classification confidence; this is unexpected, considering the attribution methods' focus.
Figure~\ref{fig:smile}-b (rightmost), however, shows that direct insertion of a slight smile can flip the classification from negative to positive, as would be expected.

\begin{figure*}[t]
    \centering
    \begin{subfigure}[b]{0.49\textwidth}
        \centering
        \includegraphics[clip,trim={13px, 14px, 10px, 0},width=\textwidth]{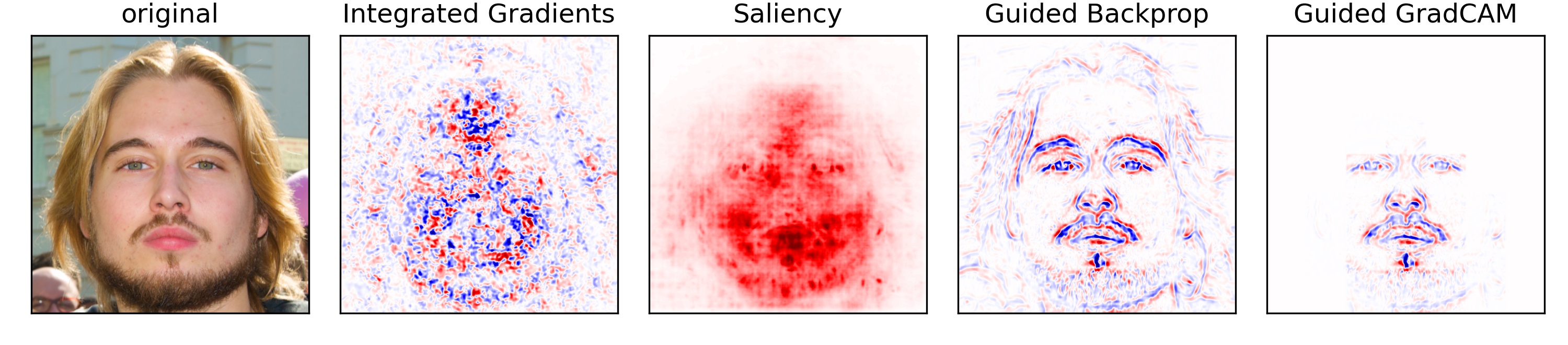}
    \end{subfigure}
    \begin{subfigure}[b]{0.49\textwidth}
        \centering
        \includegraphics[clip,trim={13px, 14px, 10px, 0},width=\textwidth]{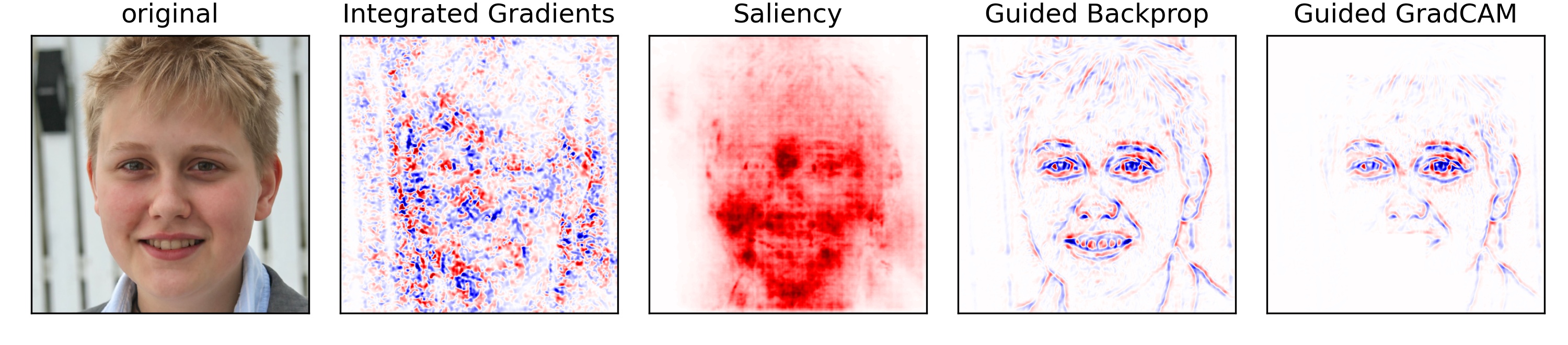}
    \end{subfigure}
    \begin{subfigure}[b]{0.49\textwidth}
    \centering
        \includegraphics[width=\textwidth]{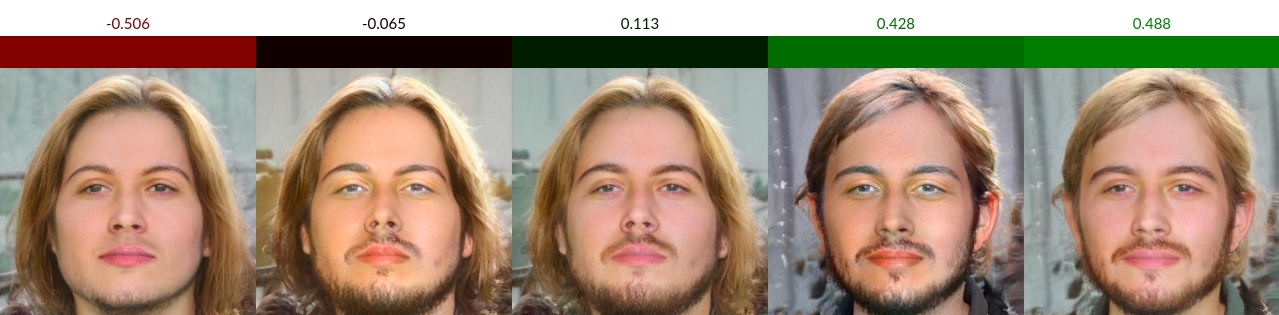}
        \caption{Individual 20299}
    \end{subfigure}
    \begin{subfigure}[b]{0.49\textwidth}
        \centering
        \includegraphics[width=\textwidth]{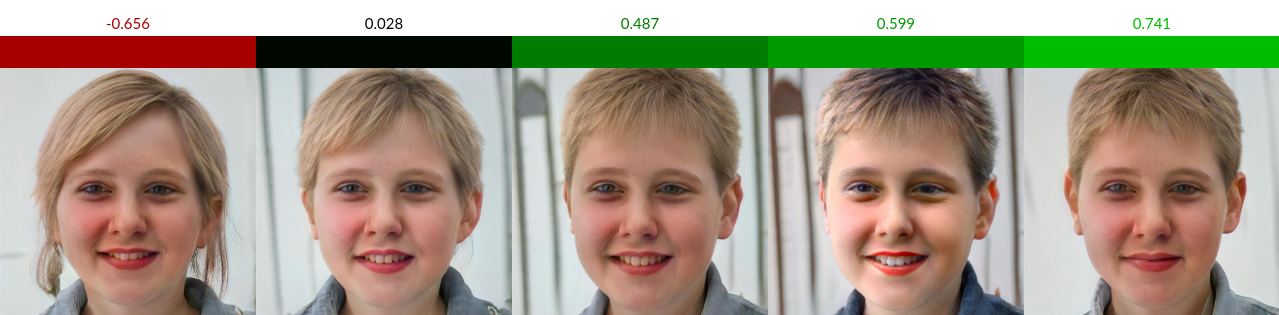}
        \caption{Individual 20937}
    \end{subfigure}
    \caption{
    Perceived Gender Classification.
    Comparison of interactive and static explanations on two different images in FFHQ.
    }
    \label{fig:gender}
\end{figure*}
Next, we investigate the effect of skin color and lighting conditions.
Naturally, the two can not always be reliably differentiated, and especially smaller variations may be due to lighting issues or skin tone differentials.
Figure~\ref{fig:teaser} shows that for the baseline image, the classification network is very uncertain; however, changing skin and hair color to darker shades (leftmost image) flips the classification to a high-confidence ``no-smile'' classification.
A caveat, in this case, is that while the critical area around the mouth seems almost unchanged except for the color, the eyes are somewhat enlarged and may also influence the decision.
This is due to the latent space of the StyleGAN2 not being perfectly disentangled.
Similar behavior happens in Figure~\ref{fig:smile}-b, where making the skin tone darker leads to a more confident negative prediction.
On the other hand, for Figure~\ref{fig:smile}-a, the darker skin tone (center image) does not change the very confident ``smile'' classification.
Increasing image brightness and skin tone in Figure~\ref{fig:smile}-c (center and rightmost) leads from a very uncertain prediction to a medium- and high-confidence ``smile'' prediction while the image in Figure~\ref{fig:smile}-d is very robust against manipulation of skin tone.

Finally, we note that the addition of makeup can also crucially change classification decisions.
This may partly be due to confounding factors in the training data set (CelebA-HQ): female celebrities at public events are more likely to both wear makeup and smile than average celebrities in other situations.
The effect is most prominent in Figure~\ref{fig:teaser}-c (rightmost).
The addition of moderate levels of makeup switches the decision from uncertain to highly confident ``smile'' without noticeable changes in the facial expression.
In Figure~\ref{fig:smile}-b (center), a similar but smaller effect can be seen, but in Figure~\ref{fig:smile}-c (rightmost), a reverse effect is visible.
This opposite trend in different images might indicate a different effect of makeup for people of different skin colors. 

Finally, we compare our method to automatically generated image modifications, similarly to \citep{denton2019image, schutte2021using}.
In Figure~\ref{fig:app-generative-static} (Appendix), we automatically modify the original reconstructed images along the following directions: ``smile,'' ``makeup,'' ``white/black''  (simplified as binary class, as provided by the FairFace model \citep{karkkainen2021fairface}), ``perceived gender'' (again, simplified and binarized; consider also the discussion in the section on perceived gender classification) and ``young/old.''
Except for the FairFace model, labels were generated using the CelebA dataset.
We emphasize that especially the concepts of perceived gender and ancestry are generally neither discernible by visual inspection alone nor binary or even categorical concepts.
We include them since they still confer information of interest for the analysis of classifiers (such as whether a classifier performs worse on images of marginalized groups) and since they illustrate a critical drawback of automated methods: they can only consider well-defined simplistic abstractions that usually fall short of reality.
See the Appendix for more details.

Notice how face morphology is considerably changed even by attributes such as ``smile'' and ``makeup'' that are presumably orthogonal to smiling.
Due to the heavy confoundedness of all features, the insights gained from the interactive approach are not apparent here.
For example, the modification of skin color is confounded by the smiling attribute in Figure~\ref{fig:app-generative-static}-a, b and c; the ``not smiling'' classification when only changing skin and hair color (in Figure~\ref{fig:teaser}-c) can not be found in the automated explanations.
Similarly, the changed amount of makeup for Figure~\ref{fig:smile}-c that leads to a no-smile classification is not visible in Figure~\ref{fig:app-generative-static}-b since both the face morphology and the facial expression along the makeup dimension change.

Importantly, this heavy confounding of latent feature dimensions also complicates the interpretation of summary statistics aggregated over many generated explanations. 
In sum, while the automatically generated explanations can yield an interesting starting point, they only provide an overly simplistic glimpse into the network's decision-making process; fine details are usually missed.
Due to space constraints, we will not discuss the automatically generated explanations in the following sections but provide all images in the code repository.

\subsection{Age estimation}
Next, we consider age estimation.
We use the FairFace model \citep{karkkainen2021fairface}, a ResNet34 trained on a diverse set of facial images.
Figure~\ref{fig:age} again compares attribution maps against snapshots from our interactive approach.
The attribution heat maps are mostly inconclusive, distributing attention over the whole image.
Arguably, for some of the images, Guided GradCAM seems to focus more on the forehead and hair (or lack thereof).

With our interactive approach, we investigated different hypotheses.
First, for the images in Figure~\ref{fig:age}-a (center) and c (leftmost and center), we removed the glasses from the faces, which appears to only have a minor effect on the estimation, with no clear direction.
Removing facial hair in Figure~\ref{fig:age}-a does not have a consistent effect, while graying out hair makes the person appear considerably older for the predictor.
Next, we investigated the influence of skin color on prediction results.
For Figure~\ref{fig:age}-c (rightmost), darkening the skin color leads to a slightly higher age prediction.
For Figure~\ref{fig:age}-b (leftmost and rightmost) and d (second), however, skin tone changes did not have a consistent effect in either direction, although skin color modification also led to a slight change in facial features.
In Figure~\ref{fig:age}-b (second) and d (center and penultimate), changes in clothing color and lighting had a noticeable effect.
To review the potential effect of scalp hair on age estimation, as indicated by Guided GradCAM, Figures~\ref{fig:age}-b (penultimate) and d (rightmost) show that more hair seems to increase the perceived age of the child considerably, but again with small, potentially confounding changes in facial features.

\subsection{Perceived Gender Classification}
Finally, we consider the classification of perceived gender, again using the FairFace model \citep{karkkainen2021fairface}, see Figure~\ref{fig:gender}.
FairFace was trained under the implicit assumptions that gender is a binary category and that gender can be accurately discerned from an image.
These assumptions are problematic and likely harmful towards marginalized groups such as transgender, intersex, and non-binary people \citep{keyes2018misgendering}.
However, we decided to include the use case since perceived gender classifiers are currently in use and are likely to be used in the future in real-world applications \citep{scheuerman2019computers}.
Consequently, we believe it is crucial to understand their inner workings.
See, e.g., \citep{denton2019image,keyes2018misgendering} for further discussion.

In both analyzed images, heat map attributions are mostly inconclusive, with some focus on the eyes, eyebrows, nose, and mouth and potentially the facial hair in Figure~\ref{fig:gender}-a.

We interactively first investigated the effect of makeup on the classification decision. In Figure~\ref{fig:gender}-a (second), the addition of makeup moved the decision only very slightly, and when makeup was applied together with shorter hair, the length of the hair dominated the effect. 
In Figure~\ref{fig:gender}-b (second to last) addition of makeup (together with a brightening of lightening conditions) moved the decision more into the direction of ``male.''
Next, we checked hairstyle and length.
In Figure~\ref{fig:gender}-a (rightmost), short hair leads to moderately-confident ``male'' prediction, even if makeup is additionally applied (second to last).
Similarly, for Figure~\ref{fig:gender}-b, longer hair (leftmost) and a slightly different hairstyle (second) move the classification in the direction of ``female.'' 
Lastly, we removed most of the facial hair in Figure~\ref{fig:gender}-a (leftmost), leading the network to change the prediction to female.
These observations may indicate that the model primarily relies on the hairstyle and amount of facial hair for its decisions.

\begin{figure}[t]
    \begin{subfigure}[b]{0.49\textwidth}
    \centering
        \includegraphics[width=\textwidth]{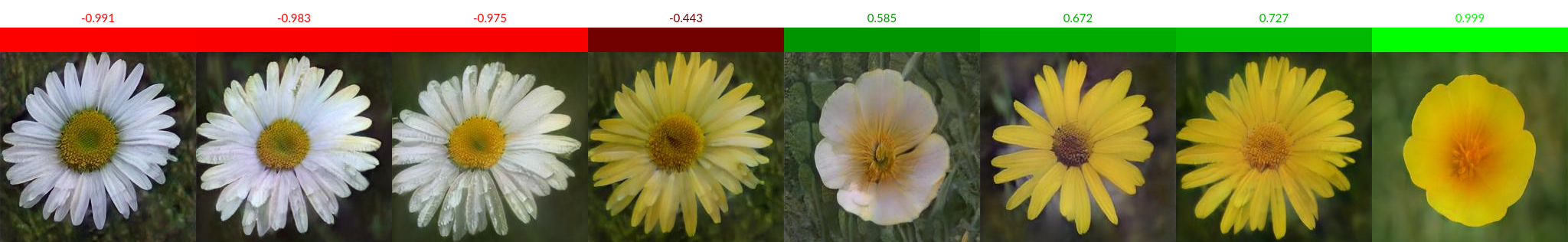}
        \caption{``Oxeye daisy'' (left) versus ``Californian poppy'' (right).}
    \end{subfigure}
    \begin{subfigure}[b]{0.49\textwidth}
        \centering
        \includegraphics[width=\textwidth]{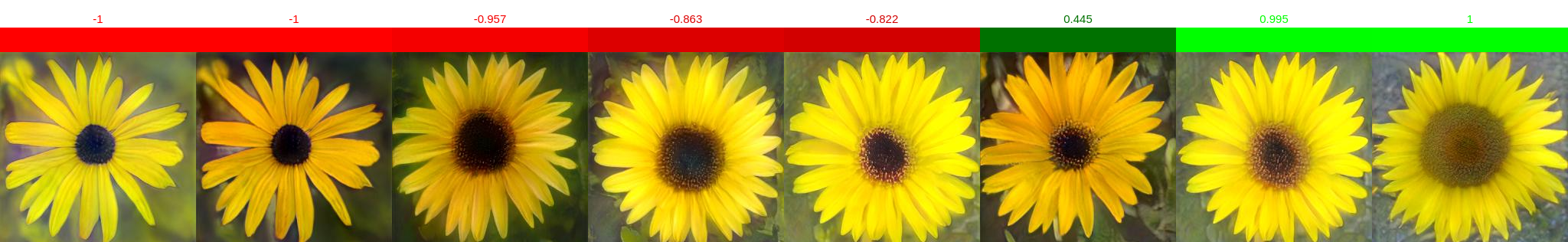}
        \caption{``Black-eyed Susan'' (left) versus ``Sunflower'' (right).}
    \end{subfigure}
  \caption{Model exploration.}
  \label{fig:flowers-exp}
\end{figure}

\subsection{Model Exploration}

We now investigate how an interactive explainability approach can also yield more general insights into a classifier's behavior.

Consider Figure~\ref{fig:flowers-exp}-a; the classifier was trained to distinguish ``oxeye daisy'' from ``Californian poppy'' from the Oxford Flowers data set \cite{nilsback2008automated}.
The main differentiator between the two classes found by the network appears to be color; this can be seen especially clearly in images 6 and 7 since their structural makeup mostly resembles the daisy flower, and only the coloring is different.
However, careful manipulation of the images shows that the network is also sensitive to several non-color features.
Namely, image 4 is classified as a daisy with moderate confidence, despite the color being closer to a Californian poppy's color pattern.
Similarly, image 5 has a color pattern similar to the color pattern of a daisy -- white petals with a yellow center -- but the flower's form is much closer to the poppy; surprisingly, the network again classifies the image as a poppy flower with moderate confidence.

Next, in Figure~\ref{fig:flowers-exp}-b, the classifier distinguishes between ``black-eyed Susan'' and ``sunflower.''
Several probes indicate that the classifier is largely invariant to the specific shade of yellow of the petals, e.g., between images 1 and 2.
However, the model reacts very sensitively towards changes in the disc florets (the center of the flower head): images 5 and 7 have almost identical petals, but image 5's disc florets are darker and slightly smaller, leading to a ``black-eyed Susan'' prediction.
Images 3 \& 4 show that a large center alone is not sufficient to classify the image as ``black-eyed Susan,'' and image 6, in contrast, shows that a small center is also no sufficient for classification as a ``sunflower''

The examples illustrate how both networks' decisions are highly complex and do not lend themselves to ``linearized'' explanations such as ``larger petals increase the likelihood of class $+1/-1$'' or similar overly simplistic abstractions.

\section{Limitations \& Conclusion}
We presented an interactive framework to explore, probe, and test a neural network's decision-making processes.

The main drawback of our proposed method is the reliance on high-quality image generation and reconstruction, particularly in the form of a StyleGAN2.
The method can be easily extended to other image synthesis and editing models.
Most state-of-the-art algorithms work very well on standardized images, as are common in facial analysis (after cropping to bounding boxes), the medical domain, and industrial production settings, but are less reliable in more open-ended applications.
High-quality image generation currently works best with a training set of at least several thousand images \citep{Karras2020ada}.
Additionally, a biased training data set will likely lead to worse image reconstructions and modifications of underrepresented groups.
Presumably, better data sets and further advances in generative models may enable the application of interactive approaches also to less standardized image applications and counteract data set biases.

Another drawback of our method is that training a high-quality image generation network may take a relatively long time and computational resources.
This investment is reasonable in high-stakes scenarios but may be overly restrictive in other cases.

Finally, we note that even in interactive approaches, confirmation bias may still persist -- the difference is that an interactive approach \emph{allows} a user to test their hypotheses.
However, data can still be misunderstood and miscommunicated. 

We believe it is prudent to draw upon a diverse set of complementary interpretability methods instead of relying on one single approach or class of approaches.
As such, we envision our interactive framework -- and extensions thereof -- as one tool in the tool belt of a practitioner or scientist.


\section*{Acknowledgements}
This project has received funding by the German Ministry of Research and Education (BMBF) in the SyReal project (project number 01|S21069A) and the German Research Foundation (DFG) project KL 2698/2-1.
MKloft acknowledges support by the Carl-Zeiss Foundation and through the DFG awards KL 2698/2-1 and KL 2698/5-1, and the BMBF awards 01IS18051A, 031B0770E, and 01IS21010C.

\bibliography{aaai22.bib} 

\appendix

\begin{table*}[]
\begin{center}
\begin{tabular}{lccc}\toprule
Classification Task & Training/Validation Set Size & Baseline & Validation Accuracy \\ 
\midrule
Smiling                                   & 22,500/7,500 & 53.0\% & 93.6\% \\
``Oxeye daisy'' vs. ``Californian poppy'' & 113/38       & 67.5\% &  100\%  \\
``Black-eyed Susan'' vs. ``Sunflower''    & 86/29        & 53.0\% & 100\% \\
\bottomrule
\end{tabular}
\end{center}
\caption{Performance of classification networks. Baseline is majority class percentage.}
\label{tab:clf}
\end{table*}

\section{Implementation Details}

Our interactive interface runs in a web app developed using Svelte/Sapper in TypeScript, accessible with most modern browsers.
It requires a backend server running the StyleGAN2 and classification network.

We used the StyleGAN2 version by \citet{karras2020analyzing,Karras2020ada} with adaptive augmentations\footnote{\url{https://github.com/NVlabs/stylegan2-ada-pytorch/}. We also provide an implementation of our interface with another open-source implementation of StyleGAN2 with a more permissive license, but especially the reconstructions were considerably better using the original implementation.}.
For the FFHQ-trained model we used the weights provided by the authors.
For the flowers dataset, we trained a new model on all images in the dataset using the default settings for $256\times 256$ pixels (\texttt{--cfg=paper256}). 

Next, we describe the training process for all classification networks used in this work (except the FairFace models, where we re-used the original authors' models).
We used ResNet18 architectures (with one-dimensional output layer for binary classification), trained with cross-entropy loss and the Adam optimizer \citep{kingma:adam}.
We fixed the learning rate to 0.0003 for all situations, used transfer learning from weights pre-trained on ImageNet, first trained for 5 (30 for flowers) epochs with frozen convolutional body and then another 5 (5 for flowers) epochs for the whole network.
For the flowers dataset, we also used random transformations (rotations by up to 180° and mirroring) to avoid overfitting due to the small sample size.
See Table~\ref{tab:clf} for classification results.
We found the results to be mostly invariant towards reasonable changes in the training procedure.

To display classification scores between $-1$ and $+1$ (instead of probabilistic scores between 0 and 1) we replaced the final sigmoid activation function after training with a hyperbolic tangent (tanh) activation function. 

FairFace's age regression model treats the setting as multi-class classification, grouping adjacent ages into categories (e.g., 0-2, 3-9, 10-19).
Our framework can integrate both ordinal regression and multi-class classification by displaying the most confident class together with the confidence score for this class (or, for few classes, displaying the confidences for each individual class).
To get a finer resolution, however, we transform the classification logits into continuous scores by averaging the center of all age brackets weighted by the softmax classification scores.

We include code to reproduce all settings in the code repository.\footnote{\url{https://github.com/HealthML/explainability-requires-interactivity}}

Most of our experiments were run on NVIDIA GeForce RTX 2080 Ti graphic cards with 11 GB memory.
However, for StyleGAN2 models with high resolution (e.g., $1024 \times 1024$ pixel) and when using many styles at the same time, we used an NVIDIA A100 40 GB GPU (the experiments required less than 16 GB at all times).

The main computational burden of the interface (besides training the networks) lies in computing the reconstructions of real images, taking up to 2 minutes per image. Image modification via style mixing is near instantaneous.

\section{Automated Generative Methods}
Here, we describe the baseline comparison method for automated generative explanations, similar to \citet{denton2019image,schutte2021using}, leveraging a linear classifier in the StyleGAN2's latent space.

\begin{figure*}
    \centering
    \begin{subfigure}[b]{0.49\textwidth}
        \centering
        \includegraphics[ width=\textwidth]{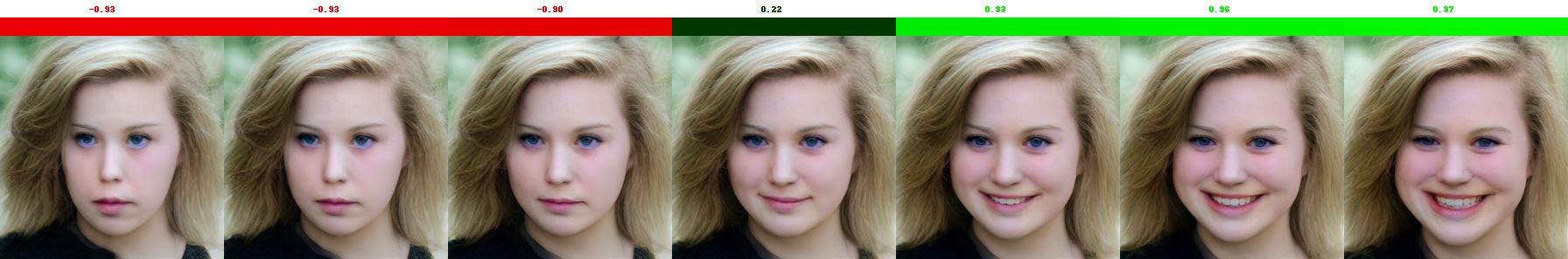}
    \end{subfigure}
    \begin{subfigure}[b]{0.49\textwidth}
        \centering
        \includegraphics[ width=\textwidth]{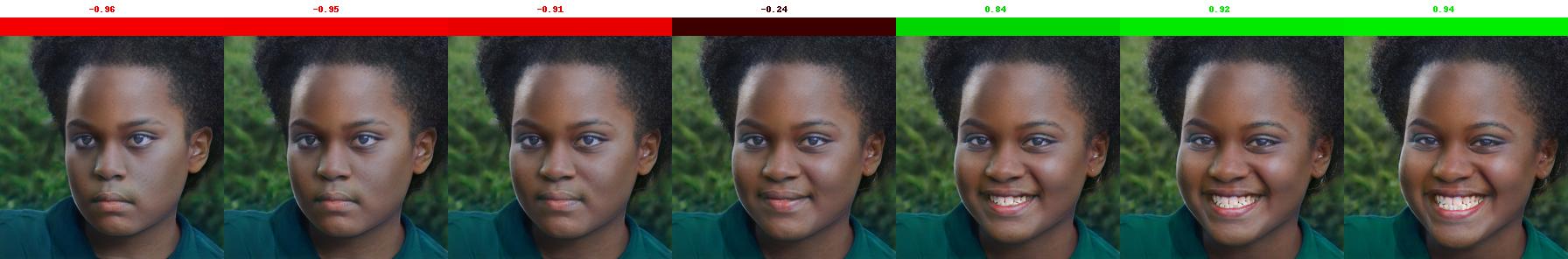}
    \end{subfigure}
    \begin{subfigure}[b]{0.49\textwidth}
        \centering
        \includegraphics[ width=\textwidth]{figures/static_generative/smile/makeup/canvas_20604.jpg}
    \end{subfigure}
    \begin{subfigure}[b]{0.49\textwidth}
        \centering
        \includegraphics[ width=\textwidth]{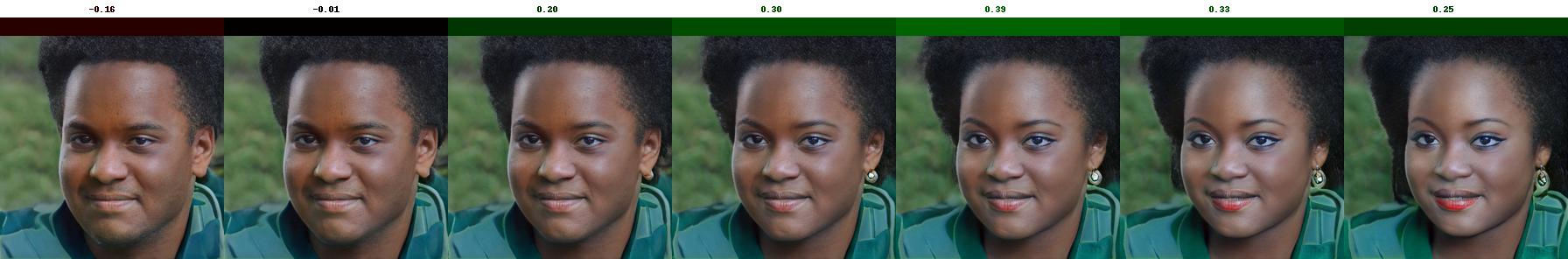}
    \end{subfigure}
    \begin{subfigure}[b]{0.49\textwidth}
        \centering
        \includegraphics[ width=\textwidth]{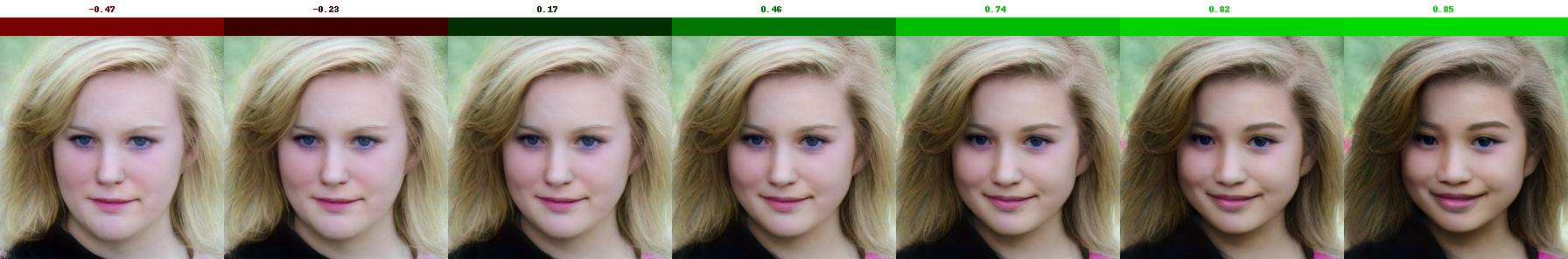}
    \end{subfigure}
    \begin{subfigure}[b]{0.49\textwidth}
        \centering
        \includegraphics[ width=\textwidth]{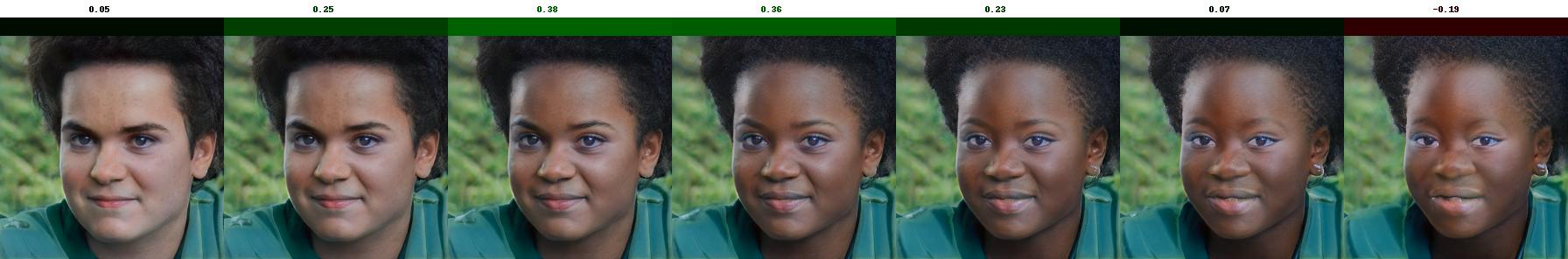}
    \end{subfigure}
    \begin{subfigure}[b]{0.49\textwidth}
        \centering
        \includegraphics[ width=\textwidth]{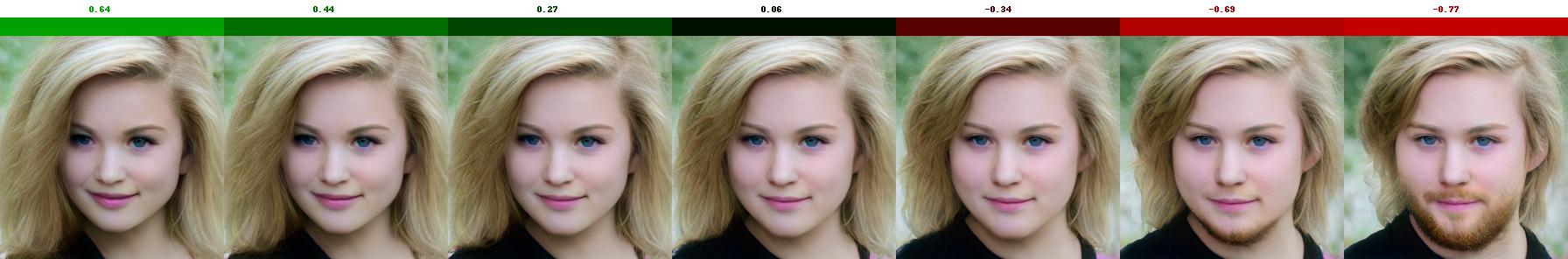}
    \end{subfigure}
    \begin{subfigure}[b]{0.49\textwidth}
        \centering
        \includegraphics[ width=\textwidth]{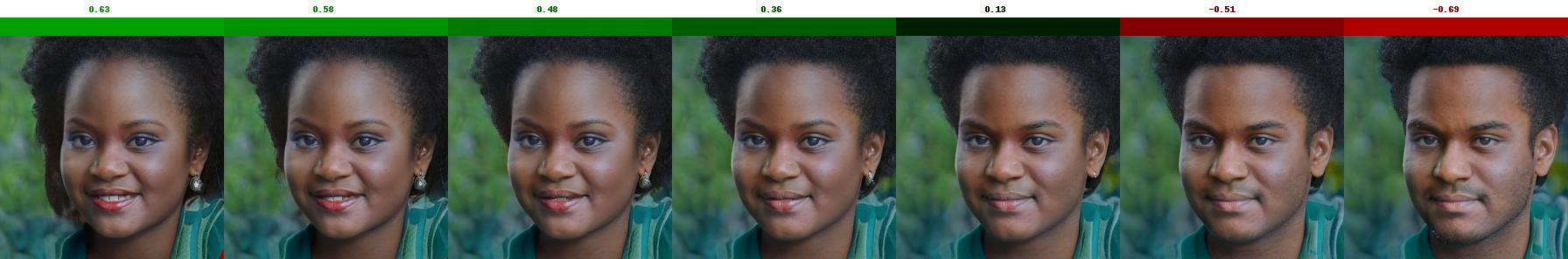}
    \end{subfigure}
    \begin{subfigure}[b]{0.49\textwidth}
        \centering
        \includegraphics[ width=\textwidth]{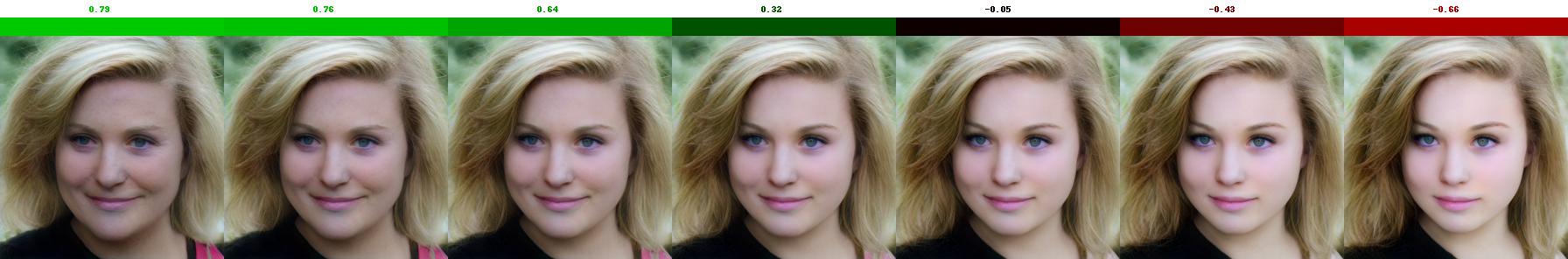}
        \caption{Individual 20604}
    \end{subfigure}
    \begin{subfigure}[b]{0.49\textwidth}
        \centering
        \includegraphics[ width=\textwidth]{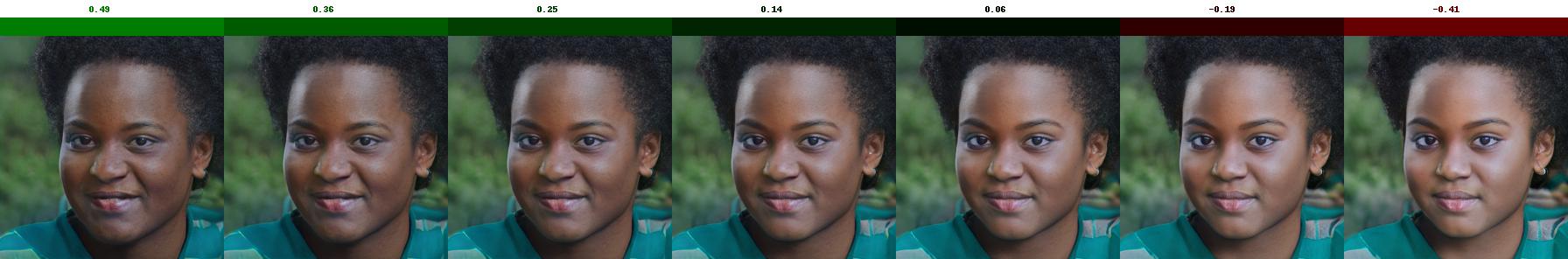}
        \caption{Individual 20659}
    \end{subfigure}

    \begin{subfigure}[b]{0.49\textwidth}
        \centering
        \includegraphics[ width=\textwidth]{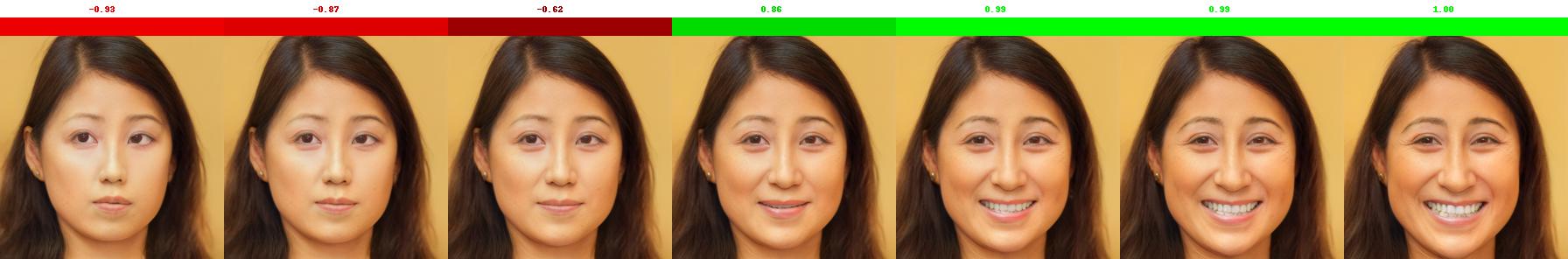}
    \end{subfigure}
    \begin{subfigure}[b]{0.49\textwidth}
        \centering
        \includegraphics[ width=\textwidth]{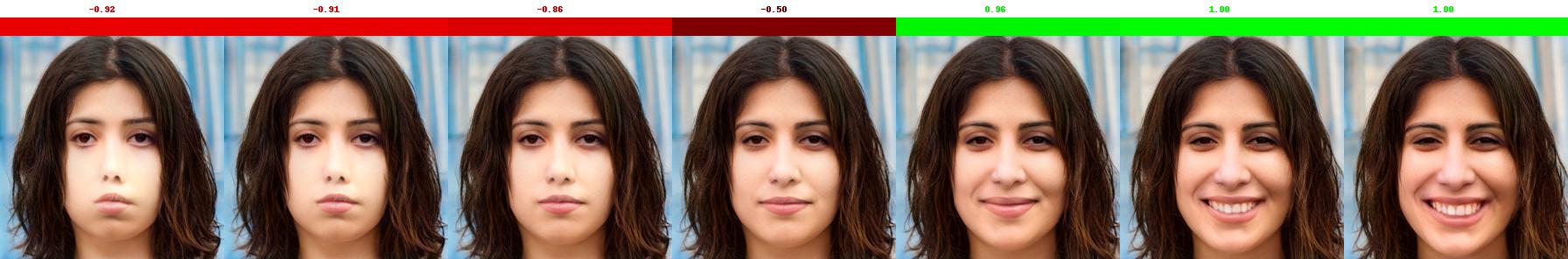}
    \end{subfigure}
    \begin{subfigure}[b]{0.49\textwidth}
        \centering
        \includegraphics[ width=\textwidth]{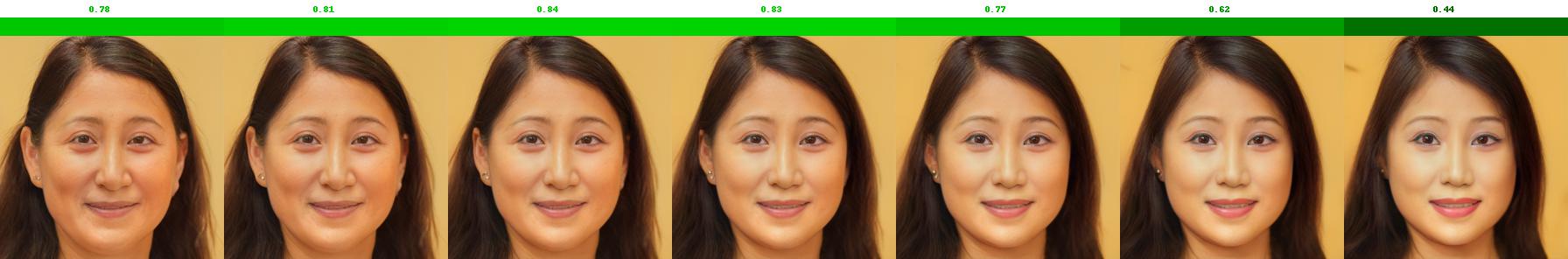}
    \end{subfigure}
    \begin{subfigure}[b]{0.49\textwidth}
        \centering
        \includegraphics[ width=\textwidth]{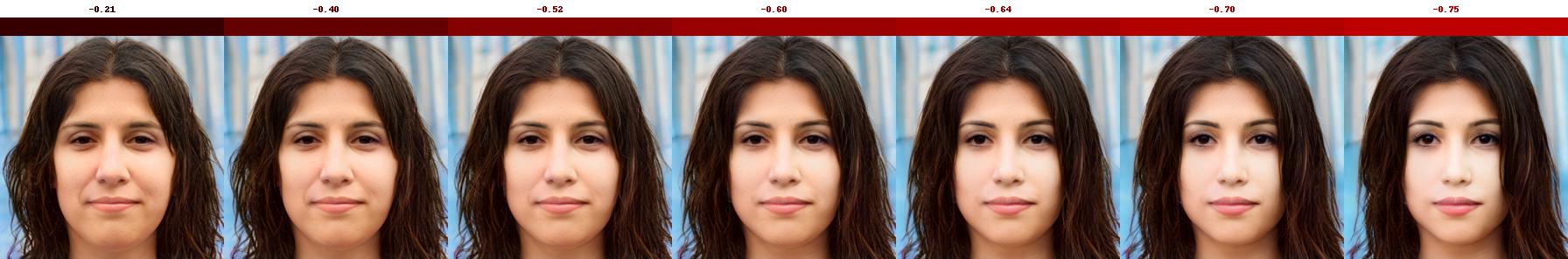}
    \end{subfigure}
    \begin{subfigure}[b]{0.49\textwidth}
        \centering
        \includegraphics[ width=\textwidth]{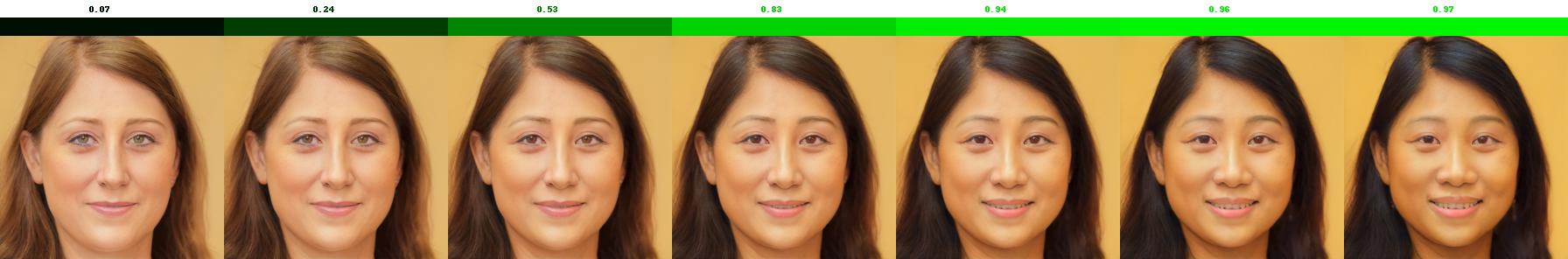}
    \end{subfigure}
    \begin{subfigure}[b]{0.49\textwidth}
        \centering
        \includegraphics[ width=\textwidth]{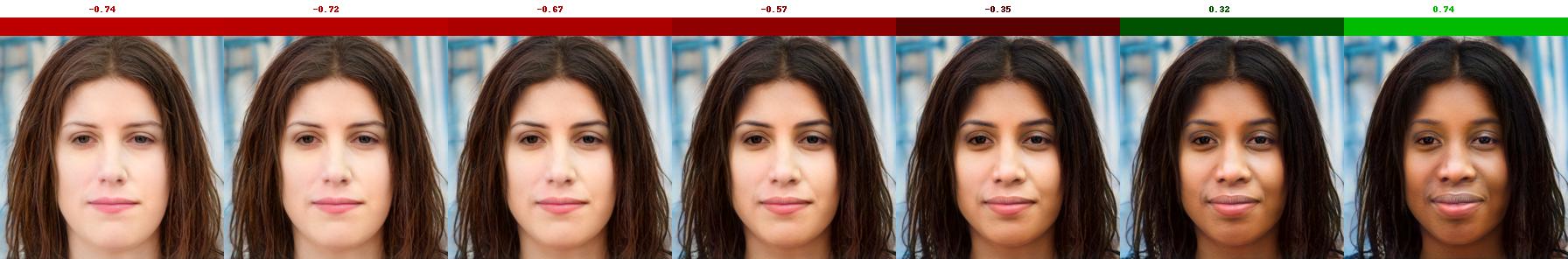}
    \end{subfigure}
    \begin{subfigure}[b]{0.49\textwidth}
        \centering
        \includegraphics[ width=\textwidth]{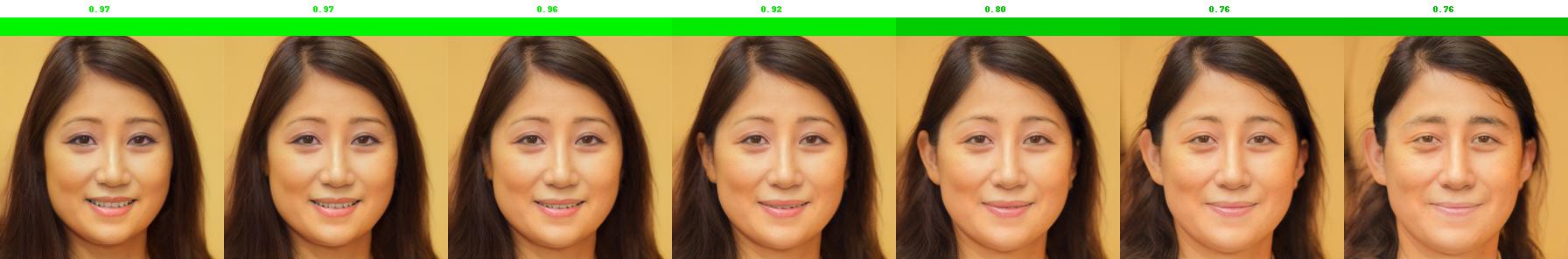}
    \end{subfigure}
    \begin{subfigure}[b]{0.49\textwidth}
        \centering
        \includegraphics[ width=\textwidth]{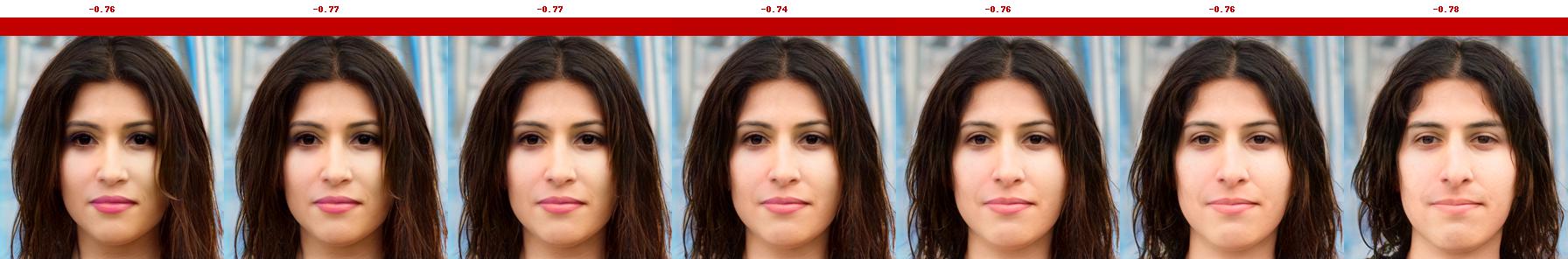}
    \end{subfigure}
    \begin{subfigure}[b]{0.49\textwidth}
        \centering
        \includegraphics[ width=\textwidth]{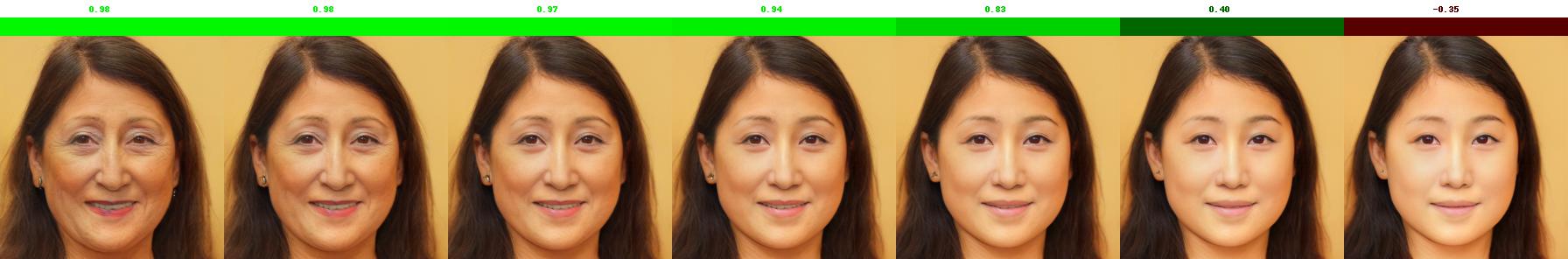}
        \caption{Individual 20000}
    \end{subfigure}
    \begin{subfigure}[b]{0.49\textwidth}
        \centering
        \includegraphics[ width=\textwidth]{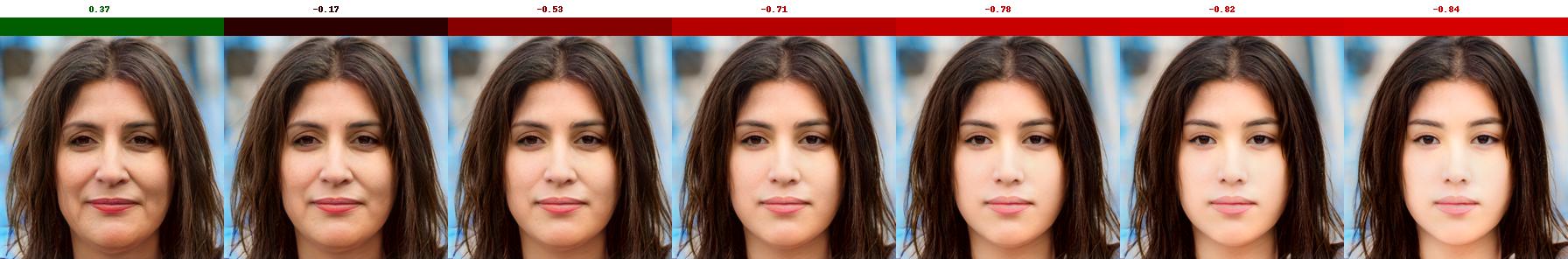}
        \caption{Individual 20827}
    \end{subfigure}
    \caption{
        Automatic generated explanations for four individuals from the FFHQ dataset.
        Each row changes the reconstructed image (center) along the following dimensions, according to a classifier in the latent StyleGAN2 space (in order from top to bottom): ``Smiling'', ``Heavy makeup'', ``Black/White'', ``Female/Male'', ``Young''.
        Note the discussion regarding classification of skin color and perceived gender in the text.
    }
    \label{fig:app-generative-static}
\end{figure*}

In a first step, we trained a ResNet18 binary classifier on images and labels from the CelebA dataset \citep{lee2020maskgan}.
We used the classes ``smiling'', ``heavy makeup'', ``male'', and ``young''.
Additionally, we used the trained FairFace model \citep{karkkainen2021fairface} only distinguishing ``black'' and ``white'' individuals (we also experimented with the ``light skin'' attribute in the CelebA dataset, but the results were far worse than the FairFace labels).
We note that these categories in the original papers were only determined by visual inspection.
Especially the labels for the ``male/female'' and the ``black/white'' categories (but also the ``young'' category) are of questionable validity.
Neither gender nor ancestry can be reliably ascertained visually, nor are those binary categories (and at least one of the individuals in Figure~\ref{fig:app-generative-static} is neither white nor black).
However, we include it here for better illustration of the potentials and drawbacks of the baseline method.
To better understand the complex interdependencies between different categories and potentially confounding factors, a more careful analysis, as demonstrated, e.g., by \citet{denton2019image}, is necessary but beyond the scope of this article.

In the next step, we randomly created a training and validation set of latent code vectors drawn from a multivariate standard normal distribution (20k in train, 5k in validation).
For each instance in training and validation set, we created a label by feeding it into the StyleGAN2 trained on FFHQ to create a synthetic image and then scoring it with the classification networks trained in the first step.
Finally, we trained a logistic regression model on these latent code vectors.

To automatically change images across each of these classifiers, we first reconstructed the image (as in the Methods Section).
Then we shifted the latent code vector along the (normalized) weight vector of the logistic regression classifier in both directions.
We found that adding $\lambda \in \{-0.09, -0.06, -0.03, 0.03, 0.06, 0.09\}$ times the logistic regression  weight vector to the reconstructed vector was usually sufficient.
Higher values often introduced strong image artifacts.

\section{Relationship to Adversarial Examples}
In the applications discussed in this work, the respective changes to the image are generally minor but realistic and perceptible.
This is in stark contrast to adversarial examples \citep{szegedy2014intriguing}, where images are modified in unrealistic (for natural images) but imperceptible ways.
Authentic variations of natural images may indicate actual biases, confounding factors, and potential failure modes in non-adversarial settings (e.g., when a practitioner has complete control over the image pipeline).
Adversarial examples, on the other hand, illustrate technical limitations and occur only in adversarial settings (e.g., if a potential attacker can modify images that go into an ML system).

\begin{figure*}[t]
    \centering
        \includegraphics[width=\textwidth]{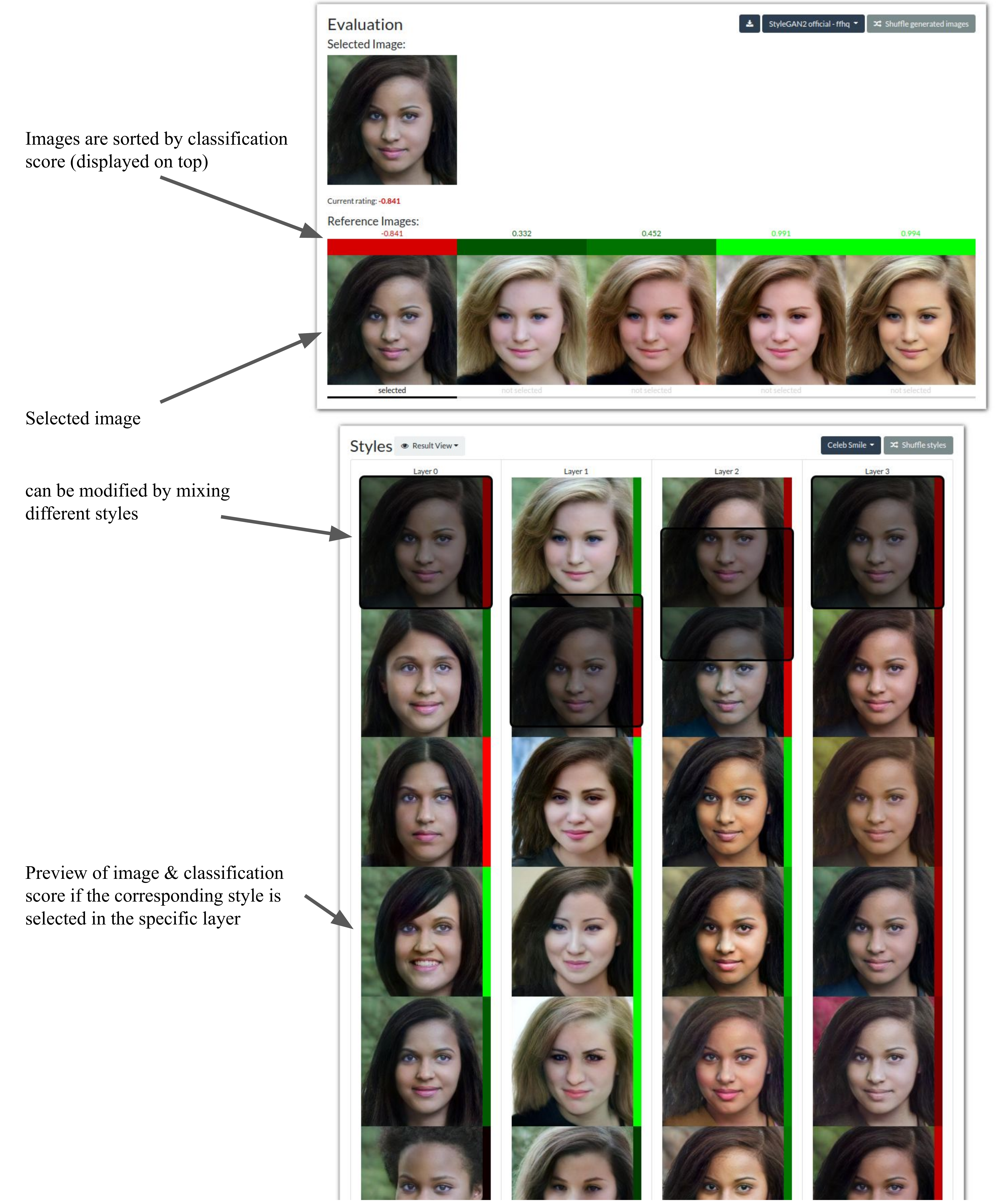}
  \caption{Overview of our interface. Images can be individually selected and modified by mixing different styles at different layers. Classification scores next to each style indicate the effect of each change. In the current view, the 18 layers of the StyleGAN2 are grouped into four layer-blocks to facilitate faster exploration.}
  \label{fig:overview}
\end{figure*}

\end{document}